\documentclass[a4paper,fleqn]{cas-dc}

\usepackage[no-math]{fontspec}
\newfontfamily\figfont{texgyretermes-regular.otf}[
  BoldFont = texgyretermes-bold.otf,
  ItalicFont = texgyretermes-italic.otf,
  BoldItalicFont = texgyretermes-bolditalic.otf
]

\usepackage[numbers,sort&compress]{natbib}
\usepackage{graphicx}
\usepackage{grffile}
\DeclareGraphicsExtensions{.pdf,.png,.jpg,.jpeg}
\usepackage{tikz}
\usetikzlibrary{arrows.meta,calc,patterns}

\usepackage{subcaption}
\usepackage{amsmath}
\usepackage{amssymb}
\usepackage{amsfonts}

\usepackage{booktabs}
\usepackage{multirow}
\usepackage{threeparttable}
\usepackage{siunitx}
\usepackage{array}
\usepackage{textcomp}
\usepackage{placeins}
\usepackage{url}
\usepackage{verbatim}
\usepackage[normalem]{ulem}
\usepackage{makecell}
\usepackage{tabularx}
\usepackage{pifont}
\usepackage{enumitem}
\usepackage{capt-of}
\usepackage{setspace}
\usepackage{algorithmic}
\usepackage{algorithm}

\usepackage{xcolor}
\usepackage{colortbl}
\definecolor{darkred}{RGB}{139,0,0}

\usepackage[protrusion=true,expansion=false,nopatch=footnote]{microtype}
\usepackage{etoolbox}
\makeatletter
\patchcmd{\@afterheading}{\clubpenalty \@M}{\clubpenalty 150}{}{}
\makeatother
\AtBeginDocument{\let\printorcid\relax}

\hypersetup{
	colorlinks=true,
	linkcolor=blue,
	citecolor=blue,
	urlcolor=magenta,
	filecolor=cyan,
	hypertexnames=false
}


\makeatletter\@secpenalty=\z@\makeatother

\renewcommand{\arraystretch}{1.12}
\newcommand{\tabcaption}[1]{\caption{\small\MakeUppercase{#1}}}
\newcommand{\tabnote}[1]{\par\vspace{2pt}{\footnotesize\raggedright\noindent #1\par}}

\AtBeginDocument{% 
}

\ExplSyntaxOn
\cs_gset_eq:NN \__movr_orig_reset_fig: \__reset_fig:
\cs_gset_eq:NN \__movr_orig_reset_tbl: \__reset_tbl:
\cs_set:Npn \__reset_fig:
  {
    \__movr_orig_reset_fig:
    \skip_set:Nn \l_fig_abovecap_skip { 4pt }
    \skip_set:Nn \l_fig_belowcap_skip { 4pt }
  }
\cs_set:Npn \__reset_tbl:
  {
    \__movr_orig_reset_tbl:
    \skip_set:Nn \l_tbl_abovecap_skip { 4pt }
    \skip_set:Nn \l_tbl_belowtbl_skip { 4pt }
  }
\ExplSyntaxOff

\graphicspath{{./}{ch1/}{ch2/}{ch3/}{ch4/}{ch5/}{ch6/}{ch7/}{images/}}

\def\tsc#1{\csdef{#1}{\textsc{\lowercase{#1}}\xspace}}
\tsc{MRSI}
\tsc{CNN}
\tsc{GNN}
\begin{document}
% Use single line spacing throughout the manuscript.
\linespread{1.0}\selectfont
\setlength{\emergencystretch}{3em}
\let\WriteBookmarks\relax
\def\floatpagepagefraction{1}
\def\textpagefraction{.001}

\shorttitle{MOVR-SAR: A Multi-resolution SAR Registration Dataset under Orbit-crossing and View-variant Imaging Differences}
\shortauthors{Yuchi Pu et al.}

\title [mode = title]{MOVR-SAR: A Multi-resolution SAR Registration Dataset under Orbit-crossing and View-variant Imaging Differences}

% --- AUTHOR DEFINITIONS ---
\author[1]{Yuchi Pu}

\author[1]{Bosong Qi}

\author[1,2]{Pengcheng Shi}

\author[1]{Jiayuan Li}[orcid=0000-0000-0000-0000]
\cormark[1]

% --- AFFILIATIONS AND FOOTNOTES ---
\affiliation[1]{organization={School of Remote Sensing and Information Engineering, Wuhan University}, city={Wuhan}, postcode={430079}, country={China}}

\affiliation[2]{organization={Hubei Luojia Laboratory},
                city={Wuhan},
                postcode={430079},
                country={China}}

\cortext[cor1]{Corresponding author. Email: ljy\_whu\_2012@whu.edu.cn.}

\csgdef{mark@corau4}{1}

\begin{abstract}
Synthetic aperture radar (SAR) image registration under varying acquisition geometries is challenging because orbit direction, look side, and incidence angle affect backscatter, terrain-induced distortion, and feature visibility. However, large-scale geometry-aware benchmarks remain scarce. This paper presents MOVR-SAR, a multi-resolution dataset of 450\,845 registered and quality-screened tile pairs derived from 2774 scene pairs from Capella, Umbra, Sentinel-1, and NISAR. It covers X, C, and L bands, ground sampling distances of 0.25--20 m, and diverse regions and land-cover types. A unified framework defines four imaging-geometry pair categories from orbit direction and look side, plus viewing-angle-separation labels for Capella and Umbra, enabling stratified evaluation under same- and opposite-direction illumination. Systematic evaluation of representative matching methods shows that official and cross-modal pretrained weights perform poorly on high-resolution X-band imagery. Fine-tuning on MOVR-SAR improves accuracy and reduces disparities across platforms, imaging-geometry pair categories, and viewing-angle separations. For example, fine-tuning increases FlexGlue AUC@5 from 3.06\% to 89.79\% on Capella and from 1.79\% to 89.03\% on Umbra, reaching 93.41\% and 94.77\% on Sentinel-1 and NISAR, respectively. Fine-tuning substantially reduces the official-weight same-orbit, same-look-side advantage and reverses the three-method average category ranking on Capella, suggesting that this bias reflects training-data coverage rather than an invariant limitation of cross-geometry matching. Land-cover effects remain platform- and method-dependent. MOVR-SAR provides a training resource and geometry-aware benchmark for robust SAR--SAR registration. The dataset and model weights will be released at \url{https://github.com/fdhzhz5870/MOVR-SAR}.\looseness=-1
\end{abstract}


\maketitle

% ---------------------------------------------------------------------
% Use single line spacing throughout, including body text and captions.
% Chapters remain in separate chN/chN.tex files and are input in order.
% ---------------------------------------------------------------------
% =====================================================================
% ch1/ch1.tex — Same-modality SAR registration under varying illumination directions
% =====================================================================

\section{Introduction}\label{sec:introduction}

Image registration increasingly encompasses multimodal data and substantial geometric and appearance changes, supported by datasets for optical--SAR, optical--map, and optical--LiDAR registration~\cite{jiang2021review}. Such benchmarks are essential for developing robust methods and enabling consistent evaluation across sensing conditions. Yet even within the same modality, SAR images of a location can differ markedly in geometry and radiometry when acquired from different viewing geometries. Because SAR provides all-day, all-weather imaging, it is indispensable for disaster response, resource exploration, environmental monitoring, building-height estimation, radargrammetry and three-dimensional reconstruction, and multitemporal change analysis~\cite{soergel2009stereo,yao2019oblique,zou2023ssta}; registration is a prerequisite for fusing such multi-resolution and multi-temporal observations, and spatial misalignment propagates errors into downstream tasks such as data fusion and change detection. This problem remains understudied and lacks dedicated data resources.

The difficulty is inherent to coherent side-looking SAR imaging. Slant-range--Doppler localization ties image geometry to the radar line of sight~\cite{moreira2013tutorial}. Terrain relief produces direction-dependent foreshortening, layover, and shadow~\cite{kropatsch1990layover}, while backscatter varies with observation geometry and surface orientation~\cite{small2011flattening}. Distinctive structures may therefore be displaced, merged, or occluded rather than merely changed in intensity. These effects are also spatially varying: a terrain patch that is compressed or brightly illuminated in one acquisition may be reversed or completely shadowed in another, as illustrated by the four imaging-geometry pair categories in Figure~\ref{fig:four_categories_concept}. Moreover, variations in satellite configuration---incidence angle, radar viewing direction, orbital altitude, and acquisition time---induce substantial geometric distortions and alter local spatial correspondences, amplifying feature-representation discrepancies; speckle and heterogeneous scattering near land-cover boundaries further degrade feature discriminability. Consequently, locally distinctive structures do not necessarily preserve their visibility, shape, or relative position across views. These structural and radiometric changes undermine appearance-based matching, especially as viewing differences increase~\cite{soergel2009stereo}.

SAR--SAR registration has advanced through improved local features~\cite{dellinger2015sarsift}, hierarchical matching for oblique images~\cite{yao2019oblique}, geometry-aware multiview registration~\cite{xiang2022geometry}, self-supervised learning~\cite{zou2023ssta}, and adaptive transformation estimation~\cite{mao2023adaptive}. However, methods developed for a specific sensor or acquisition setup may perform well within a limited viewing regime yet fail to generalize across orbit directions and look sides. A key bottleneck is the availability of SAR--SAR data: dedicated datasets are rare, and the few public ones are small and narrow in coverage. More fundamentally, a large-scale, multi-resolution dataset that systematically stratifies samples by imaging-geometry pair category and viewing-angle separation remains lacking. The scarcity of geometry-aware annotations obscures how acquisition conditions affect matching performance and limits both reliable benchmarking and targeted learning.

To fill this gap, this paper introduces \textbf{MOVR-SAR} (Multi-resolution Orbit-crossing View-variant SAR Registration), comprising approximately 450\,000 registered and quality-screened SAR--SAR tile pairs from Capella, Umbra, Sentinel-1, and NISAR. Its unified geometry taxonomy and quality-controlled coarse-to-fine construction enable explicit modeling, stratified evaluation, and targeted learning of acquisition-induced differences (Figure~\ref{fig:Overview of the MOVR-SAR}). This paper makes three main contributions:

\begin{figure*}[!tb]
    \centering
    \includegraphics[width=0.95\textwidth]{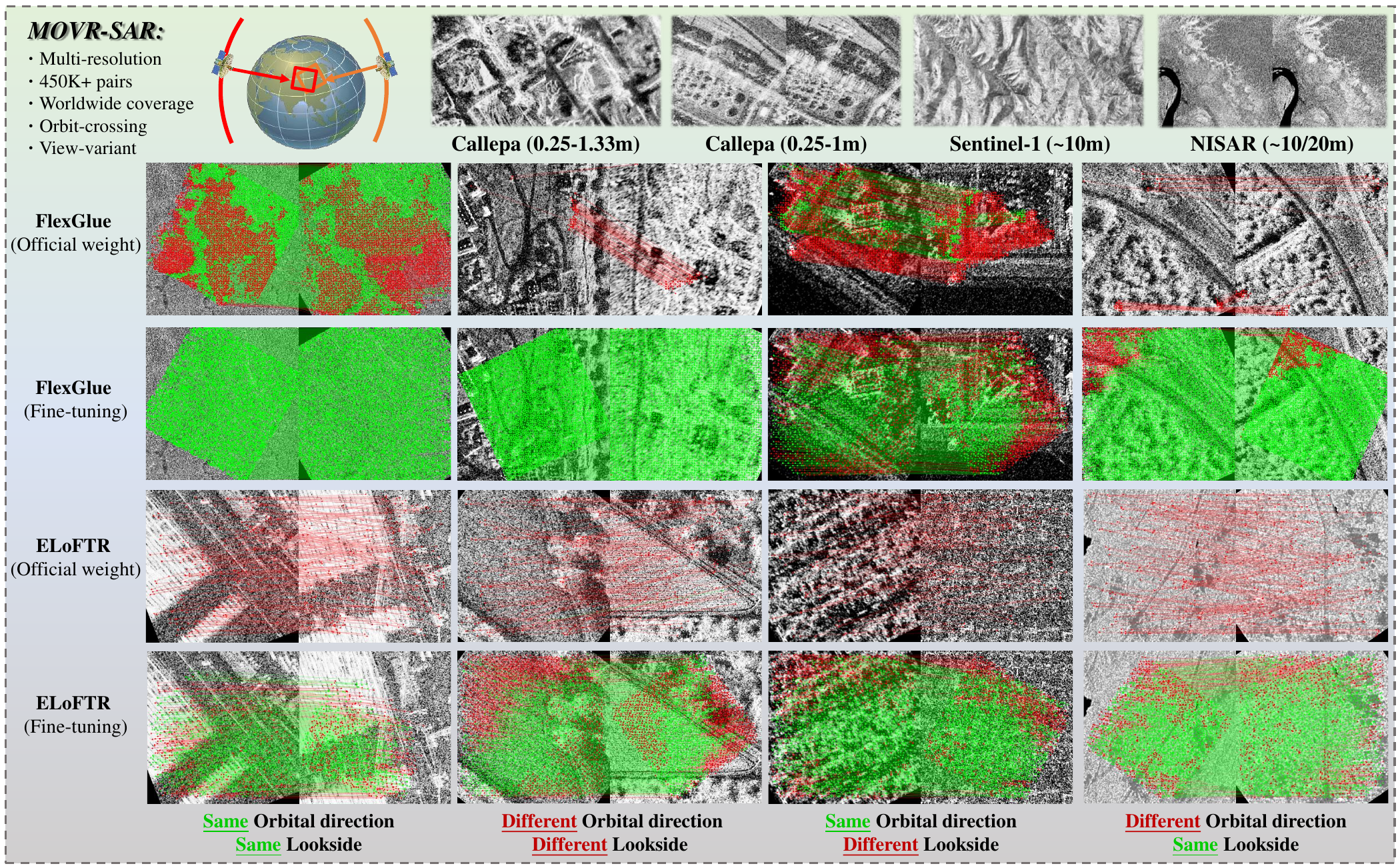}
    \caption{Overview of the MOVR-SAR dataset. The left panel lists its key properties (multi-resolution, 450\,000+ pairs, worldwide coverage, orbit-crossing, and view-variant imaging), while the examples present representative SAR tile pairs of the four imaging-geometry pair categories (OSLS, ODLD, OSLD, and ODLS) from Capella, Umbra, Sentinel-1, and NISAR, together with qualitative matching results of FlexGlue~\cite{lindenberger2026flexglue} and ELoFTR~\cite{wang2024eloftr} using official versus fine-tuned weights.\looseness=-1}
    \label{fig:Overview of the MOVR-SAR}
\end{figure*}

\textbullet\ We propose a unified framework for characterizing SAR imaging-geometry differences. It combines four metadata-based imaging-geometry pair categories with exact and approximate angular measures of line-of-sight separation, providing a basis for dataset organization and stratified evaluation.

\textbullet\ We construct MOVR-SAR, comprising approximately 450\,000 registered, quality-screened tile pairs from Capella, Umbra, Sentinel-1, and NISAR, spanning three frequency bands and ground sampling distances of 0.25--20 m. Each pair is annotated with its imaging-geometry pair category and dominant land-cover class, supporting geometry-aware evaluation and targeted learning.

\textbullet\ A systematic evaluation of 10 matching methods in 23 configurations compares pretrained and fine-tuned models across data sources, imaging-geometry pair categories, and viewing-angle separations. The results show that fine-tuning on MOVR-SAR substantially reduces geometry- and angle-dependent performance gaps.

% =====================================================================
% ch2/ch2.tex
% Section 2: Related Work
% =====================================================================

\section{Related Work}\label{sec:related_work}

This section reviews related work on public SAR datasets, image registration methods, and dedicated SAR--SAR registration, and identifies the gap that MOVR-SAR is designed to fill. Together, these strands motivate the geometric framework, dataset, and benchmark developed in the following sections.

\subsection{SAR Imagery Datasets}
\label{sec:sar_datasets}

Public SAR datasets have mainly supported image interpretation rather than registration. HRSID provides high-resolution SAR imagery with ship annotations for detection and segmentation~\cite{wei2020hrsid}, OpenSARUrban contains labeled Sentinel-1 patches for urban scene classification~\cite{zhao2020opensarurban}, and SARDet-100K provides a larger multi-category benchmark for SAR object detection~\cite{li2024sardet100k}. Their labels describe targets or semantic classes, not spatial correspondences between SAR acquisitions. More importantly, these datasets do not organize multiple acquisitions of the same region into pixel-wise aligned SAR imagery and are not specifically designed for registration. Open SAR archives provide broader scene coverage, but public datasets for SAR image matching are dominated by cross-modal (SAR--optical) resources, whereas dedicated SAR--SAR datasets remain scarce and comparatively restricted in public availability. To our knowledge, no dedicated public dataset currently supports systematic training and evaluation of SAR--SAR registration across platforms and viewing geometries, which motivates~MOVR-SAR.

\subsection{Image Registration Methods}
\label{sec:reg_methods}

Existing registration methods can be divided into handcrafted and learning-based approaches. Handcrafted methods typically follow a detect--describe--match pipeline. SIFT detects scale-invariant keypoints and represents them with gradient histograms, providing a general solution for moderate scale and viewpoint changes~\cite{Lowe2004}. SAR-SIFT adapts this framework to speckle through ratio-based gradients and a SAR-specific scale space~\cite{dellinger2015sarsift}, while RIFT uses phase congruency to reduce sensitivity to nonlinear radiometric differences~\cite{li2020rift}. These methods require no task-specific training, but their reliability depends on the repeatability and visibility of local structures across~acquisitions.

Learning-based methods replace one or more handcrafted components with data-driven models. SuperPoint learns keypoints and descriptors in a self-supervised manner, and SuperGlue uses a graph neural network to reason about assignments and unmatched points~\cite{detone2018superpoint,sarlin2020superglue}. LightGlue improves this framework through adaptive computation~\cite{lindenberger2023lightglue}, MambaGlue through state-space modeling~\cite{ryoo2025mambaglue}, MapGlue targets multimodal remote-sensing matching~\cite{wu2026mapglue}, and HIMO proposes cross-modality invariant feature transform~\cite{gao2026himo}. Detector-free methods instead establish correspondences without explicit keypoints: LoFTR introduces Transformer-based coarse-to-fine matching~\cite{sun2021loftr}, ELoFTR improves its efficiency~\cite{wang2024eloftr}, FlexGlue supports sparse-to-dense matching~\cite{lindenberger2026flexglue}, GeoSAM incorporates geometry-aware guidance~\cite{zhang2026geosam}, and RoMa v2 predicts dense correspondence fields~\cite{edstedt2025romav2}. Together, these methods cover sparse, semi-dense, and dense matching.

Cross-modal methods further address appearance differences between sensing modalities. XoFTR adapts detector-free matching to heterogeneous pairs~\cite{tuzcuouglu2024xoftr}, MINIMA learns modality-invariant matching through large-scale multimodal training~\cite{ren2025minima}, and MatchAnything pursues universal cross-modal matching with large-scale pre-training~\cite{he2025matchanything}. Nevertheless, general optical or cross-modal pretraining does not explicitly model view-dependent SAR distortions, making dedicated evaluation and adaptation necessary for SAR--SAR registration.

\subsection{SAR--SAR Registration}
\label{sec:sarsar_registration}

Research dedicated to SAR--SAR registration has progressed from geometry-aware handcrafted matching to self-supervised learning. For difficult viewing conditions, Yao et al. combine complementary filtering with multilevel matching for oblique SAR imagery~\cite{yao2019oblique}, and Xiang et al. introduce imaging-geometry constraints for multiview high-resolution SAR registration~\cite{xiang2022geometry}. Learning-based methods reduce reliance on manually designed features and labels: SARPointNet learns descriptors from automatically generated samples~\cite{li2022sarpointnet}, SSTA-Net combines SAR-Superpoint with transformation aggregation for self-supervised multitemporal registration~\cite{zou2023ssta}, and Mao et al. adapt alignment transformations through iterative self-supervised optimization~\cite{mao2023adaptive}. In practice, however, general-purpose image registration methods remain the common choice, whereas dedicated SAR--SAR methods are comparatively few and none of the cited works provides a public implementation. These studies are also typically evaluated on independently constructed datasets for specific sensors and regions rather than on a shared public benchmark, which has hindered consistent cross-platform comparison and practical adoption.

% =====================================================================
% ch3/ch3.tex
% Section 3: Geometric characterization of SAR imaging differences
% =====================================================================

\section{Geometric Characterization of SAR Imaging Differences}\label{sec:sar_imaging_difference}

SAR image pairs from different viewing geometries differ not only in radiometry but in the underlying relation between ground and image coordinates, making their registration harder than appearance-only matching. This section develops a geometric framework to characterize these differences, from side-looking imaging geometry and terrain-induced distortions to a parameterization by orbit direction, look side, and incidence angle, and finally a taxonomy of image pairs and the viewing-angle separation used throughout this paper. The framework provides a consistent basis for dataset construction and stratified~evaluation.

\subsection{Side-Looking Imaging Geometry}

SAR forms images by transmitting microwave pulses and recording surface backscatter. Range localization follows from the echo delay, which determines the target--antenna slant range
\begin{equation}
    R = \frac{c_0 \tau}{2},
\end{equation}
where $R$ is the slant range, $c_0$ is the speed of light, and $\tau$ is the round-trip propagation time. At nadir, a constant-slant-range ring produces left--right ambiguity; side-looking geometry avoids this ambiguity by directing the antenna beam away from nadir\cite{moreira2013tutorial,curlander1991synthetic,elachi1987physics}. In azimuth, platform motion forms a synthetic aperture whose coherent echo integration yields fine resolution. Under broadside (zero-squint) conditions, an antenna aperture of length $L_a$ gives an approximate azimuth resolution of $\lambda R / (2 L_a)$\cite{cumming2005digital,curlander1991synthetic}, where $\lambda$ is the carrier wavelength and $R$ is the slant range. For range bandwidth $B_r$, the slant-range resolution $\delta_r$ and its ground-range projection $\delta_{gr}$ are
\begin{equation}
    \delta_r = \frac{c_0}{2B_r}, \qquad
    \delta_{gr} = \frac{\delta_r}{\sin\theta},
\end{equation}
where $\theta$ is the incidence angle. Hence $\delta_{gr}$ depends on $\theta$. Even without terrain distortion, incidence-angle differences alter compression in slant-range--ground-range mapping and may change apparent radiometry through acquisition geometry and surface conditions\cite{moreira2013tutorial,cumming2005digital,elachi1987physics,small2011flattening}.

Unlike optical imagery, raw SAR data are sampled in slant-range--azimuth coordinates rather than projected vertically onto the ground. Terrain can therefore be displaced, compressed, reversed in range order, or occluded. Image-pair differences thus encode look-direction geometry as well as speckle and temporal change; a later subsection formalizes same-direction and opposite-direction~configurations.

\subsection{Terrain-Induced Geometric Distortions}

Consider a radar-facing slope with horizontal length $L$ and slope angle $\psi>0$ measured from the horizontal. The incidence angle $\theta$ is measured from the local vertical to the radar line of sight. Their relation defines three distortion~regimes.

\emph{Foreshortening} occurs when $0<\psi<\theta$. Under the local plane-wave approximation, the along-slope length $L/\cos\psi$ projects onto slant range as
\begin{equation}
    L_{\mathrm{slant}} = L\frac{\sin(\theta-\psi)}{\cos\psi}.
\end{equation}
For flat terrain ($\psi=0$), this reduces to $L\sin\theta$. As $\psi$ approaches $\theta$, $L_{\mathrm{slant}}\to 0$, compressing the slope into a narrow near-range interval.

\emph{Layover} occurs when $\psi>\theta$. The crest is closer to the radar than the toe, so its echo arrives first and reverses the range order; overlapping returns may produce a bright band.

\emph{Radar shadow} occurs on slopes facing away from the radar. For back-slope angle $\psi_b>0$, if
\begin{equation}
    \psi_b > 90^\circ-\theta,
\end{equation}
foreground terrain blocks illumination, leaving the region beyond it in shadow.

\begin{figure}[pos=htbp]
    \centering
    \captionsetup{font=normalsize}
    \includegraphics[width=\columnwidth]{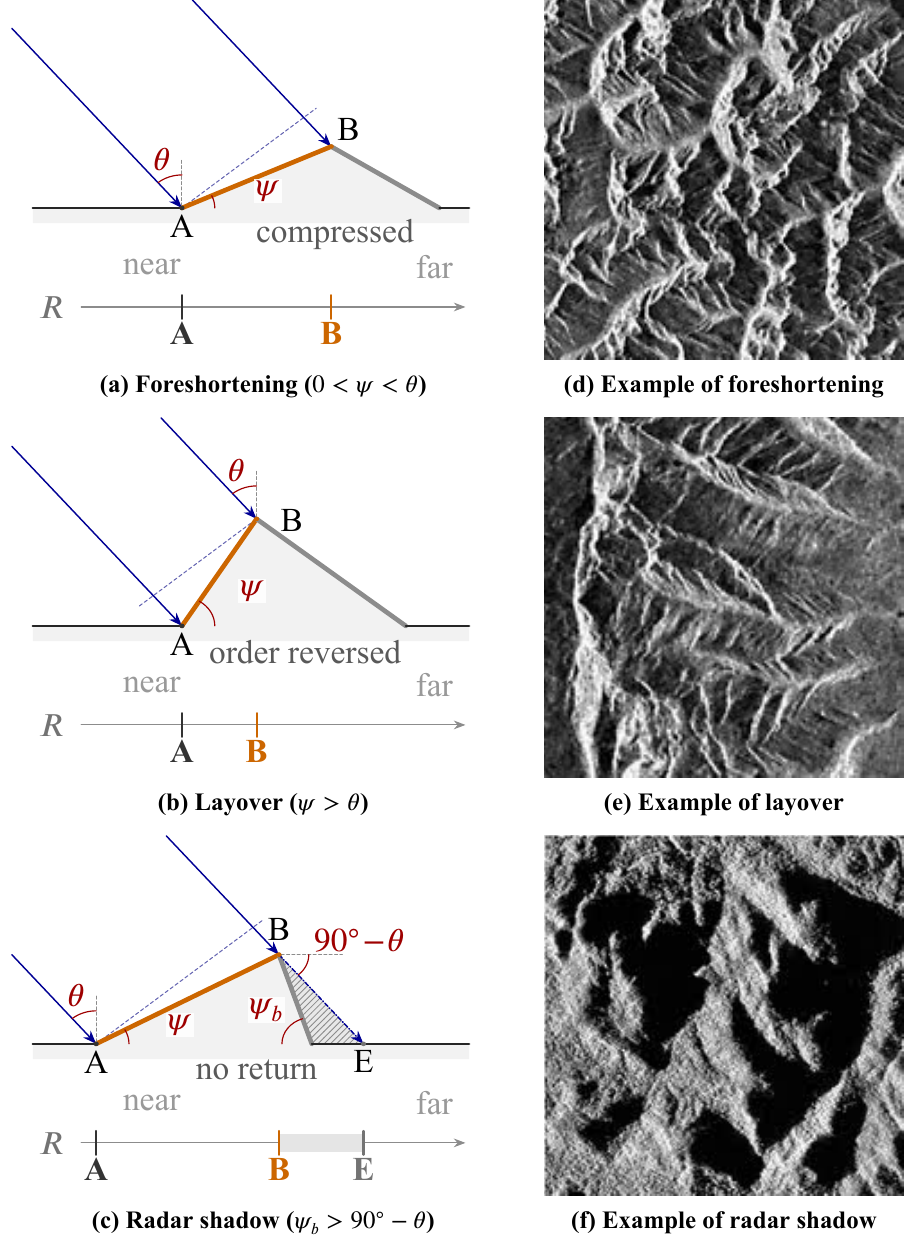}
    \caption{Schematic diagrams and corresponding SAR examples of terrain-induced distortions: (a,d) foreshortening, (b,e) layover, and (c,f) radar shadow. Blue arrows and dashed lines denote radar look directions and equal-slant-range wavefronts, respectively. In the schematics, $\theta=40^\circ$ and $\Delta R=\lvert AB\rvert\sin(\theta-\psi)$. The three effects compress range spacing, reverse range order, and produce a no-return interval ending at $E$, respectively.}
    \label{fig:distortion_theory_real}
\end{figure}

Figure~\ref{fig:distortion_theory_real} links the three distortion mechanisms to their SAR signatures: (a,d) foreshortening, (b,e) layover, and (c,f) radar shadow. Foreshortening and layover affect radar-facing slopes, whereas shadow affects slopes facing away. These criteria are stated for a two-dimensional range cross-section in which the slope faces directly toward or away from the radar; for a slope of general azimuth, the relevant quantity is its projection onto the range direction, and the local incidence angle then differs from the flat-terrain $\theta$. Illumination-direction changes thus alter backscatter and the spatial patterns of compression, overlap, and occlusion, producing geometric differences that cannot be explained by speckle or temporal change alone\cite{elachi1987physics,kropatsch1990layover}. Because acquisitions view terrain from different positions, the same feature may occupy different distortion regimes across images, limiting the effectiveness of simple rescaling. The next section relates these effects to orbit direction, look side, and incidence angle, particularly their role in reversing radar-facing and shadowed slopes.

\subsection{Parameterization of Observation Geometry}
\label{ssub:acquisition_parameters}

Acquisition geometry is characterized by orbit direction, look side, and incidence angle, which describe platform heading, left- or right-looking configuration, and line-of-sight inclination, respectively. Although speckle, temporal change, and scene properties preclude a complete sample-level model, these variables support geometry-driven grouping and the four image-pair~categories.

For platform heading $\alpha$ and horizontal look azimuth $\phi$, both measured clockwise from north, the look azimuth is
\begin{equation}
\phi = \alpha \pm 90^\circ,
    \label{eq:phi}
\end{equation}
where the signs denote right- and left-looking configurations, respectively. Here, A/D denote ascending/descending passes and R/L denote right-/left-looking acquisitions. For near-polar sun-synchronous orbits at low and middle latitudes, ascending and descending headings approach $0^\circ$ and $180^\circ$, so the four combinations AR, DL, AL, and DR yield look azimuths of approximately $\pm 90^\circ$. Changing both orbit direction and look side preserves $\phi$, whereas changing either reverses it, providing the basis for image-pair classification.

Equivalently, in a local east--north--up frame, the unit vector from a ground point toward the radar is
\begin{equation}
    \mathbf{L}(\phi,\theta)=
    \begin{pmatrix}
        \sin\phi\sin\theta\\[2pt]
        \cos\phi\sin\theta\\[2pt]
        \cos\theta
    \end{pmatrix},
    \label{eq:L}
\end{equation}
and the line-of-sight angular separation between two acquisitions is
\begin{equation}
    \Omega = \arccos\!\left(\mathbf{L}_1^\mathsf{T}\mathbf{L}_2\right),
    \label{eq:omega}
\end{equation}
where $\mathbf{L}_i=\mathbf{L}(\phi_i,\theta_i)$ is the unit look vector of acquisition $i\in\{1,2\}$. Thus, $\cos\Omega=\sin\theta_1\sin\theta_2\cos\Delta\phi+\cos\theta_1\cos\theta_2$, with $\Delta\phi=\phi_2-\phi_1$ denoting the difference between the two horizontal look azimuths. For $\Delta\phi\approx 0^\circ$ and $180^\circ$, $\Omega$ is approximately $|\theta_1-\theta_2|$ and $\theta_1+\theta_2$, respectively. These relations describe acquisition geometry, whereas pixel-area correction and radiometric terrain correction to the terrain-flattened backscatter coefficient $\gamma^{0}$ address radiometric terrain effects\cite{frey2013pixel_area,small2011flattening}.

\subsection{Imaging-Geometry Pair Categories}
\label{ssub:taxonomy}

Using the near-polar approximation, SAR image pairs are classified according to two binary metadata attributes: whether the orbit direction is the same and whether the look side is the same. The resulting four combinations are mutually exclusive and exhaustive within this representation: OSLS (same orbit, same look side), ODLD (different orbit, different look side), ODLS (different orbit, same look side), and OSLD (same orbit, different look side). By the look-azimuth relation above, OSLS and ODLD share approximately the same horizontal look direction (same-direction illumination), whereas OSLD and ODLS have approximately opposite directions (opposite-direction illumination); the four categories thus form a finer grouping beneath these two illumination-direction regimes. The taxonomy captures dominant geometric differences without claiming to provide a complete physical model of appearance variation.

Under the near-polar, low- and middle-latitude approximation, $\Delta\phi$ is restricted to the limiting cases of $0^\circ$ and $180^\circ$. OSLS and ODLD therefore share $\Omega\approx\lvert\theta_1-\theta_2\rvert$, while OSLD and ODLS share $\Omega\approx\theta_1+\theta_2$. Retaining four categories preserves orbit-direction and look-side information that is not encoded by $\Omega$ alone; pairs with similar $\Omega$ may therefore still differ in image orientation and local appearance. Category membership does not by itself determine local layover or shadow, which also depend on terrain and the actual look vectors. Moreover, a shared limiting expression does not imply identical $\Omega$ values, which depend on the headings and incidence angles of each pair. When the near-polar assumption fails or $\Delta\phi$ lies substantially between these limits, $\Omega$ should be computed directly using Eq.~\eqref{eq:omega}. The four-category classification should therefore be interpreted as a metadata-based geometric grouping rather than a continuous measure of imaging difference.

Table~\ref{tab:four_categories} summarizes the geometric definition, illumination relation, and representative angular measure of each category.

\begin{table}[!htbp]
    \centering
    \tabcaption{Geometric definition and illumination relation of the four imaging-geometry pair categories.}
    \label{tab:four_categories}
    \small
    \setlength{\tabcolsep}{3pt}
    \resizebox{\columnwidth}{!}{%
    \begin{tabular}{cccc}
        \toprule
        Pair category & Typical pair & Illumination relation & Dominant angular difference \\
        \midrule
        OSLS & AR+AR & Same direction & $\lvert\theta_1-\theta_2\rvert$ \\
        ODLD & AR+DL & Same direction & $\lvert\theta_1-\theta_2\rvert$ \\
        ODLS & AR+DR & Opposite direction & $\theta_1+\theta_2$ \\
        OSLD & AR+AL & Opposite direction & $\theta_1+\theta_2$ \\
        \bottomrule
    \end{tabular}}
    \tabnote{\textbf{Naming convention.} \textbf{S}\,=\,same and \textbf{D}\,=\,different for orbit direction and look side, respectively. \textbf{OSLS}: same orbit, same look side; \textbf{ODLD}: different orbit, different look side; \textbf{ODLS}: different orbit, same look side; \textbf{OSLD}: same orbit, different look side.}
\end{table}

\begin{figure*}[!tb]
    \centering
    \includegraphics[width=\textwidth]{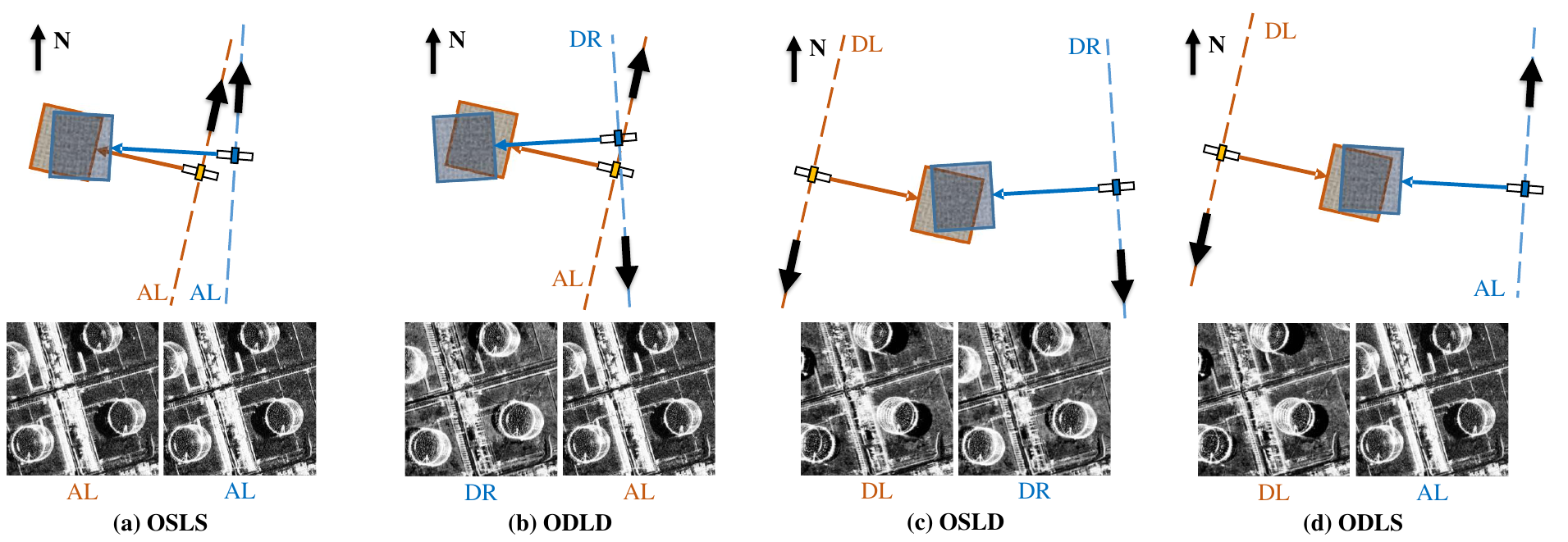}
    \caption{Schematic illustration of the four imaging-geometry pair categories and representative SAR tile pairs: (a) OSLS, (b) ODLD, (c) OSLD, and (d) ODLS. OSLS and ODLD correspond to same-direction illumination, whereas ODLS and OSLD correspond to opposite-direction illumination. A/D denote ascending/descending orbit directions, and R/L denote right-/left-looking acquisitions.}
    \label{fig:four_categories_concept}
\end{figure*}

Figure~\ref{fig:four_categories_concept} depicts the four imaging-geometry pair categories. For example, an AR+AR pair belongs to OSLS, whereas an AR+DL pair belongs to ODLD. Within pairs belonging to either category, the two acquisitions have approximately the same horizontal look azimuth, leading to similar brightness layouts and distortion locations. Under the near-polar approximation, their residual geometric differences are primarily characterized by $\lvert\theta_1-\theta_2\rvert$.

For OSLD and ODLS, representative pairs are AR+AL and AR+DR, respectively. These pairs have approximately opposite horizontal look azimuths. A terrain feature facing the radar in one image may therefore face away in the other. As illustrated above, strong returns associated with foreshortening or layover may be replaced by weak backscatter or shadow under the opposite viewing direction. For these configurations, $\theta_1+\theta_2$ is more informative than $\lvert\theta_1-\theta_2\rvert$ as a measure of the dominant geometric difference.

To summarize each pair with a single angle, the viewing-angle separation $\Delta\theta$ is defined as an approximation to $\Omega$ at the same ground location. Under the near-polar approximation, it is estimated as
\begin{equation}
\Delta\theta \;\approx\;
\left\{
\begin{array}{ll}
\left| \theta_1-\theta_2 \right|, & \text{same-direction},\\[6pt]
\theta_1+\theta_2, & \text{opposite-direction}.
\end{array}
\right.
\label{eq:dtheta}
\end{equation}
This estimator assumes nearly identical or opposite horizontal look azimuths; the approximation error depends primarily on the deviation of the actual headings from these limits. It is computed from nominal, scene-level incidence angles rather than local, pixel-wise values, and it merges two physically distinct effects, the incidence-angle difference and the near-$180^\circ$ azimuth flip, into a single category-dependent scalar. It therefore serves as a stratification measure for dataset organization and evaluation rather than a uniform pixel-level model of imaging difference. The imaging-geometry pair category is encoded as \textrm{O\{S,D\}L\{S,D\}}, where the two symbols indicate agreement in orbit direction and look side, respectively. If per-scene incidence angles are available, $\Delta\theta$ is appended as a suffix. For instance, \textrm{ODLD0.3} denotes a different-orbit, different-side pair with $\Delta\theta\approx\ang{0.3}$, while \textrm{OSLD73.5} denotes a same-orbit, different-side pair with $\Delta\theta\approx\ang{73.5}$.

The calculation of $\Delta\theta$ requires per-scene incidence angles. Among the four data sources, these angles are available in the metadata used here only for Capella and Umbra. Sentinel-1 and NISAR acquire right-looking images, so their imaging-geometry pair categories follow directly from the orbit-direction relation. Since their scheduling metadata lack incidence angles, these pairs are labeled only by the orbit flag, without a $\Delta\theta$ suffix.

These conventions are applied consistently in the dataset construction and experiments that follow.

% =====================================================================
% ch4/ch4.tex
% Section 4: MOVR-SAR Dataset
% =====================================================================

\section{MOVR-SAR Dataset}
\label{ch:dataset}

This section introduces MOVR-SAR, a multi-resolution SAR image-pair benchmark for registration under orbit-crossing and view-variant imaging differences. We then describe its data sources, coarse-to-fine construction and quality control, composition and distributions, and principal characteristics, with corresponding tiles sharing file names, aligned on a common grid, and encoding the imaging geometry and land-cover class in their paths.

%---------------------------------------------------------------------
\subsection{Data Sources}
\label{sec:sources}

The four sources are complementary in frequency band, spatial resolution, and product level, spanning representative configurations of current spaceborne SAR. For each platform, the observation parameters, selected product, and spatial reference are summarized.

\textbf{Capella}\cite{capella2024open_data} provides single-polarization X-band imagery acquired in spotlight (SP), sliding spotlight (SS), and stripmap (SM) modes. We use its geocoded terrain-corrected (GEO) products, which are orthorectified and provide initial correspondences on a common geographic grid.

\textbf{Umbra}\cite{umbra2024open_data} provides single-polarization X-band spotlight imagery. Its geocoded ellipsoid-corrected (GEC) products are used; they are projected onto the World Geodetic System 1984 (WGS 84) reference ellipsoid and carry geographic references but are not terrain-corrected using a digital elevation model (DEM). Their geolocation is used only for initialization, followed by content-based coarse registration.

\textbf{Sentinel-1}\cite{torres2012sentinel1,microsoft2026s1rtc} provides C-band interferometric wide-swath (IW), VV-polarized imagery on a 10 m grid. We use the analysis-ready radiometrically terrain-corrected (RTC) products from Microsoft Planetary Computer, which provide $\gamma^0$ backscatter and support initial correspondences on a common geographic grid.

\textbf{NISAR}\cite{rosen2017nisar,asf2026nisargcov} provides L-band, frequency A, HH-polarized imagery. The provisional Level-2 geocoded covariance (GCOV) products from NASA's Alaska Satellite Facility (ASF) are used. Their terrain-corrected, geocoded $\gamma^0$ measurements are provided on north-oriented grids with 10 m or 20 m spacing and support initial correspondences.

%---------------------------------------------------------------------
\subsection{Dataset Construction Pipeline}

\begin{figure*}[!tb]
  \centering
  \includegraphics[width=0.95\textwidth]{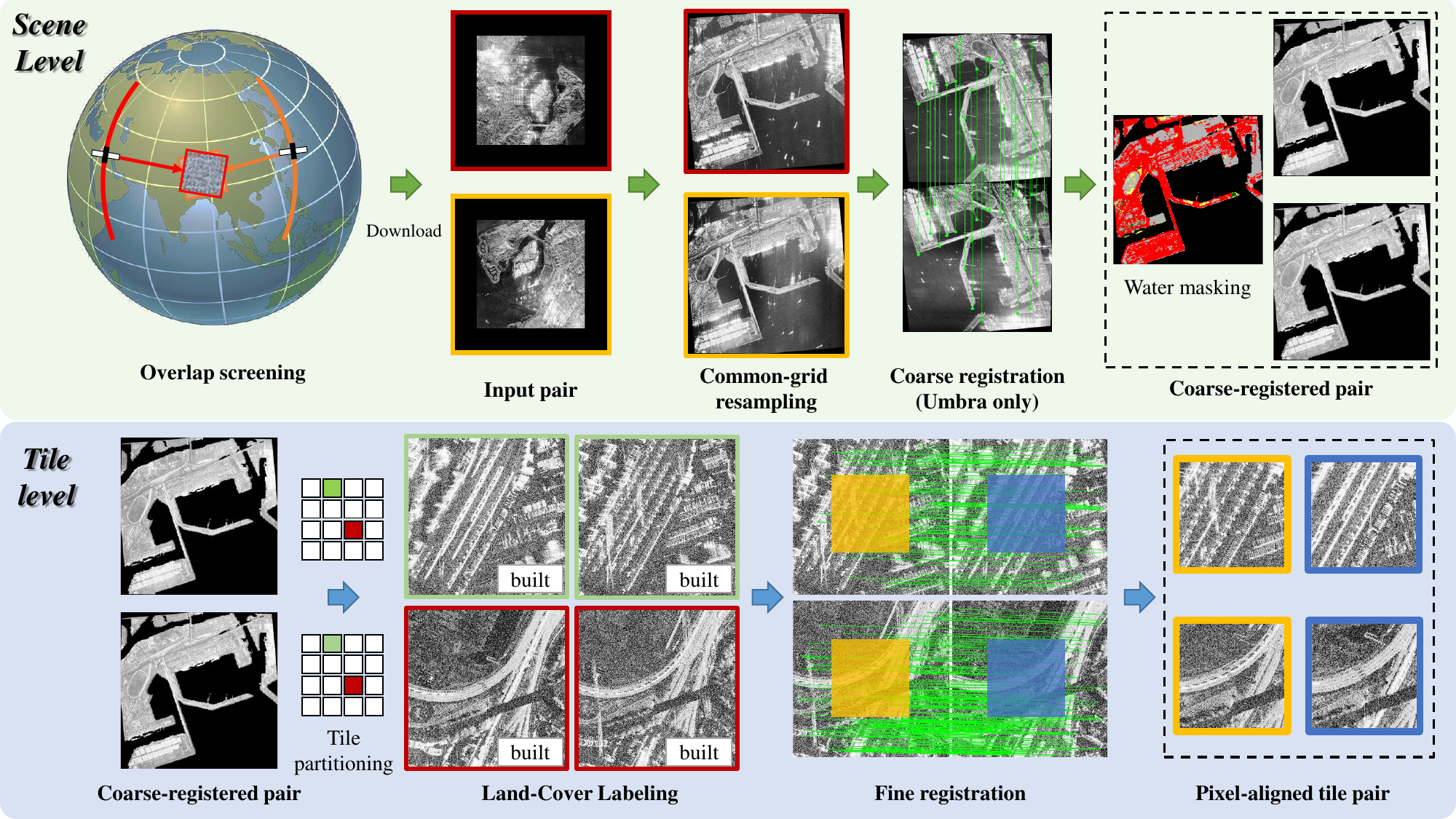}
  \caption{MOVR-SAR Composition and Distribution. Construction pipeline at the scene level and the tile level. Archive scenes are first selected by footprint overlap and then resampled onto a common grid; coarse registration is required for Umbra only, after which water masking yields the coarse-registered pair. The coarse-registered pair is cropped at the tiling stride into coarse-aligned tile pairs, which are assigned land-cover labels and then fine-registered and center-cropped into pixel-aligned tiles. The frames at the ends show each level's input and output; intermediate outputs before fine registration are termed coarse-registered images. All subsets share the same internal structure and file-naming convention.}
  \label{fig:pipeline}
\end{figure*}

\label{sec:pipeline}

MOVR-SAR is constructed to balance sample scale and geometric consistency: the former supports statistical evaluation, whereas the latter determines the reliability of the registration labels. Figure~\ref{fig:pipeline} illustrates the common coarse-to-fine pipeline used for all four subsets, comprising pair screening and sampling, coarse registration and quality screening, radiometric processing and water masking, tile partitioning and land-cover labeling, and fine registration with center cropping. The workflow is shared across subsets, while sampling rules, coarse-registration procedures, and masking requirements are adapted to each archive. The orbit direction (ascending/descending) and look side (left/right) of each acquisition are taken from the product metadata and used to assign every pair to one of the four imaging-geometry pair categories. Nevertheless, cross-subset performance differences may also reflect resolution, land-cover composition, acquisition time, and sampling distribution, and therefore cannot be attributed solely to sensor characteristics or imaging geometry.

%---------------------------------------------------------------------
\newpage
\subsubsection{Pair Screening and Sampling}
\label{sec:overlap}

An image pair must have sufficient common coverage to yield tile pairs. The scene bounding boxes are therefore prescreened using platform-specific overlap thresholds. For Capella and Umbra, the intersection of the two bounding boxes must cover at least 20\% of the smaller scene footprint. For Sentinel-1 and NISAR, the intersection-over-union (IoU) must be at least 60\%. This rapid filter removes pairs with insufficient overlap while targeting enough area for at least 20 valid tiles per scene pair once tiles are cut at the $\geq90\%$ valid-pixel threshold described later.

For Capella and Umbra, pairwise intersections are computed to enumerate unique overlap combinations. Within each imaging-geometry pair category, each scene contributes to at most one image pair, thereby reducing near-duplicate samples caused by repeated use of the same~acquisition.

Sentinel-1 and NISAR combine urban sampling with global random sampling. Urban sampling uses the world's 100 largest cities by area as candidate locations, with quotas of 25 ascending--ascending pairs, 25 descending--descending pairs, and 50 ascending--descending pairs. A pair is retained only when built-up areas account for at least 80\% of its overlap. Additional locations are sampled uniformly across continents to supplement non-urban land-cover types and bring the sizes of these two subsets to a scale comparable with Capella and Umbra.

Water prescreening generally requires the center of the overlap to lie on land. Port cities are not immediately excluded when this center lies near or within water; instead, another image pair is selected, and the city is discarded only if no suitable pair is available. This candidate-level check is later complemented by the pixel-level water masking described later.

%---------------------------------------------------------------------
\subsubsection{Coarse Registration and Quality Screening}
\label{sec:coreg}

The footprint intersection of each pair is projected into the local Universal Transverse Mercator (UTM) zone. A common grid is defined at the coarser of the two pixel spacings and aligned to the coarser scene, and the higher-resolution scene is bilinearly resampled onto this grid to provide a common spatial reference for coarse~registration.

%---------------------------------------------------------------------
\begin{table*}[!t]
  \centering
  \tabcaption{Product specifications and tile-pair composition of the four released subsets.}
  \label{tab:archives}
  \footnotesize
  \setlength{\tabcolsep}{2pt}
  \resizebox{\linewidth}{!}{%
  \begin{tabular}{l l c c c c c c r}
  \toprule
  \multirow{2}{*}{Subset} & \multirow{2}{*}{Archive / Product} & \multirow{2}{*}{Band ($\lambda$, cm)} & \multirow{2}{*}{Pixel spacing (m)} & \multicolumn{4}{c}{Imaging-geometry pair category (tile pairs and percentage)} & \multirow{2}{*}{Tile pairs} \\
  \cmidrule(lr){5-8}
   & & & & OSLS & ODLD & OSLD & ODLS & \\
  \midrule
  Capella & GEO (SP/SS/SM) & X (3.1) & 0.25--1.33 & 51\,848 (46.8\%) & 17\,384 (15.7\%) & 22\,525 (20.3\%) & 19\,143 (17.3\%) & 110\,900 \\
  Umbra & GEC (SP) & X (3.1) & 0.25--1.00 & 41\,560 (35.0\%) & 30\,831 (26.0\%) & 29\,318 (24.7\%) & 16\,872 (14.2\%) & 118\,581 \\
  Sentinel-1 & RTC (IW) & C (5.6) & 10 & 75\,920 (67.7\%) & -- & -- & 36\,285 (32.3\%) & 112\,205 \\
  NISAR & GCOV (frequency A) & L (23.6) & 10/20 & 65\,746 (60.2\%) & -- & -- & 43\,413 (39.8\%) & 109\,159 \\
  \midrule
  Total & \multicolumn{2}{l}{} & & 235\,074 (52.1\%) & 48\,215 (10.7\%) & 51\,843 (11.5\%) & 115\,713 (25.7\%) & 450\,845 \\
  \bottomrule
  \end{tabular}}
  \tabnote{A tile pair consists of two registered tiles at the same location. Pixel spacing denotes the nominal ground sampling distance of the source products. Numbers in parentheses are percentages. Abbreviations: SP, spotlight; SS, sliding spotlight; SM, stripmap; IW, interferometric wide swath; VV, vertical transmit--vertical receive; HH, horizontal transmit--horizontal receive.}
  \end{table*}

For Capella, Sentinel-1, and NISAR, the common coverage of the geocoded products is cropped directly to obtain coarse-registered pairs. For Umbra, the GEC product provides only an initial window, and coarse registration relies on content matching. An ELoFTR\cite{wang2024eloftr} model fine-tuned on Capella tiles is used, because Capella and Umbra share the X band and have comparable resolutions and spotlight imaging modes. Both scenes are downsampled for block-wise matching, and ELoFTR correspondences are used to fit an affine transformation; manual control points are selected when this procedure fails. The registered pair is then cropped to its common coverage. This content-based step only provides rough alignment for Umbra, after which a common fine-registration step is applied uniformly to all~subsets.

Manual quality screening is then applied to all coarse-registered pairs; pairs with visible quality defects or inconsistent feature layouts (unstable correspondences) are discarded. The retained counts after all screening stages are summarized in Table~\ref{tab:archives}.

%---------------------------------------------------------------------
\subsubsection{Radiometric Processing and Water Masking}
\label{sec:radiometry}

To reduce differences in calibration, quantization, and dynamic range among the data sources, global histogram equalization is applied independently to each scene after coarse registration. This operation makes contrast and input dynamic range more comparable without imposing a common physical radiometric calibration. Because only the global intensity distribution is normalized, relative, local, and textural differences induced by acquisition geometry remain in the tiles and are retained for matching.

Although pair screening requires the overlap center to lie on land, substantial water bodies may remain within a scene. Pixel-level water masking therefore focuses on oceans and large lakes using the European Space Agency (ESA) WorldCover 2021 (v200) product\cite{zanaga2022worldcover}, which provides global land-cover classes at 10 m resolution. Class 80 (permanent water bodies) is extracted, and water patches belonging to connected components of at least 2500 m$^2$ are masked. The mask is computed jointly over the union of the valid extents of the two scenes and applied identically to both, ensuring the same validity pattern by construction. For Sentinel-1 and NISAR, pixels without land-cover labels are also removed. A pixel is therefore considered valid if it is retained by the shared water mask and, for Sentinel-1 and NISAR, carries a land-cover label. After processing, water accounts for no more than 0.05\% of the joint valid area.

%---------------------------------------------------------------------
\subsubsection{Tile Partitioning and Land-Cover Labeling}
\label{sec:tiling}

After radiometric processing and water masking, the common coverage of each pair is partitioned into nonoverlapping 1024\,\texttimes\,1024-pixel tiles, with the stride equal to the tile size. Only tiles for which valid pixels account for at least 90\% of each scene are retained and saved as single-channel 8-bit PNG images.

Land-cover labels are derived from ESA WorldCover 2021\cite{zanaga2022worldcover}. The classification grid is reprojected onto the tile grid, and class pixel counts are computed over the joint valid area. The class with the largest count is assigned as the dominant tile label, and the area proportion of every class is recorded.

%---------------------------------------------------------------------
\subsubsection{Fine Registration and Center Cropping}
\label{sec:refine}

After geocoding or coarse registration, corresponding pixels remain only approximately aligned. Archive geolocation accuracy, terrain relief, and cross-geometry viewpoint differences may leave translation, rotation, and scale residuals, particularly for sub-meter X-band pairs, motivating an additional fine-registration step.

Correspondences are extracted and a homography between the two tiles is estimated using MapGlue\cite{wu2026mapglue}, a multimodal feature matcher. With the first scene of each pair taken as the reference, the moving tile is then resampled into the reference-tile coordinate system. Reprojection residuals of the estimated mapping are used to screen the tile pairs, and only those whose residuals satisfy a one-pixel threshold are retained. This metric controls the registration consistency of the released tiles but is not equivalent to absolute localization accuracy validated using independent external checkpoints. Nor does it establish strict physical pixel-wise correspondence in layover or shadow regions, where a unique one-to-one mapping may not exist.

The released tiles are the central 512\,\texttimes\,512 regions of the two registered images, reducing edge voids caused by resampling. Center cropping is not intended to remove layover, shadow, or other local view-induced distortions.

%---------------------------------------------------------------------

\subsection{Dataset Composition and Distribution}
\label{sec:composition}

\subsubsection{Scale and Spatiotemporal Distribution}

\label{sec:volume}

\begin{figure*}[!tb]
  \centering
  \includegraphics[width=0.872\textwidth]{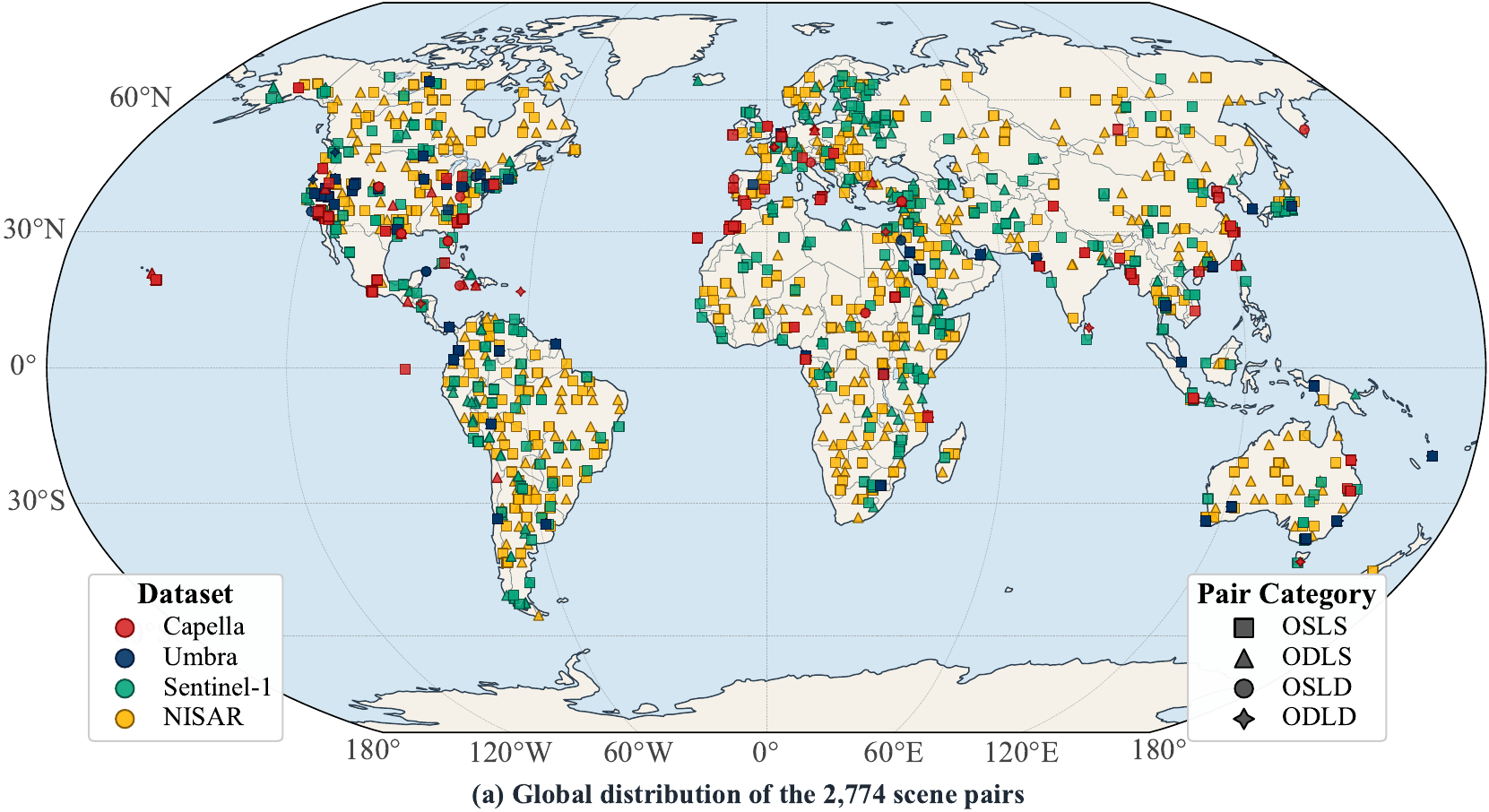}\\[2pt]
  \includegraphics[width=\textwidth]{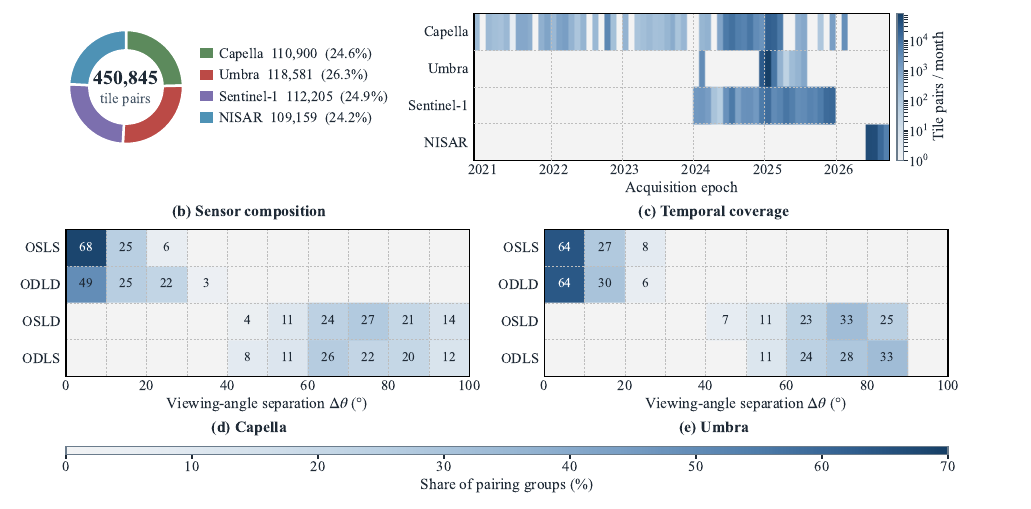}
  \caption{Geographic distribution and acquisition statistics of MOVR-SAR. (a) Global distribution of the 2774 scene pairs in Robinson projection; colors denote subsets and marker shapes denote the four imaging-geometry pair categories. (b) Tile-pair proportions of the four subsets. (c) Monthly tile-pair counts for each subset on a logarithmic color scale. (d,e) Distributions of the viewing-angle separation $\Delta\theta$ for Capella and Umbra, respectively, grouped by imaging-geometry pair category in $10^\circ$ bins. Color intensity denotes the percentage of scene pairs within each category.}
  \label{fig:worldangle}
\end{figure*}

MOVR-SAR comprises 2774 scene pairs, 4116 unique scenes, and 450\,845 tile pairs. Because a scene may be reused across imaging-geometry pair categories (though at most once within a category), the number of unique scenes is smaller than twice the number of scene pairs. Capella, Umbra, Sentinel-1, and NISAR contain 542, 1085, 523, and 624 scene pairs, respectively, and 110\,900, 118\,581, 112\,205, and 109\,159 tile pairs. Table~\ref{tab:archives} summarizes these subset sizes alongside their product specifications and imaging-geometry composition. Figure~\ref{fig:worldangle}a maps the global distribution of the scene pairs; panels b and c summarize the tile-pair proportions and monthly distributions of the four subsets.

\begin{figure*}[!tb]
  \centering
  \includegraphics[width=0.8\textwidth]{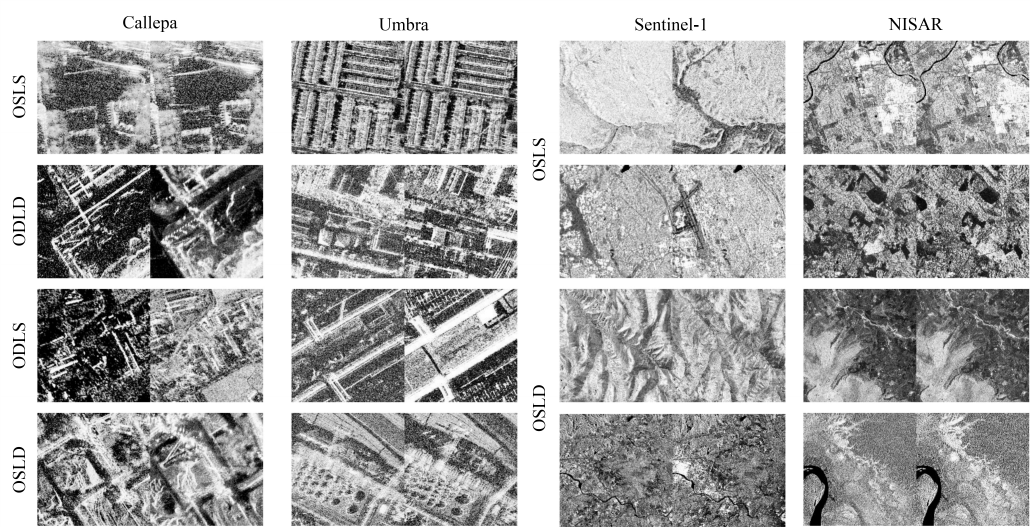}
  \caption{Representative registered SAR tile pairs from the released tiles. Rows correspond to the four imaging-geometry pair categories and columns to the four subsets; each cell shows the two registered images side by side. Differences in acquisition geometry, radiometry, and acquisition time remain visible after registration.}
  \label{fig:examples}
\end{figure*}

Capella and Umbra cover all four imaging-geometry pair categories. For Capella, OSLS, ODLD, OSLD, and ODLS contain 227, 87, 140, and 88 scene pairs, respectively; the corresponding counts for Umbra are 354, 291, 261, and 179. Sentinel-1 and NISAR are right-looking and contain only OSLS and ODLS. These two categories comprise 276 and 247 scene pairs in Sentinel-1, and 313 and 311 in NISAR. Across the full dataset, OSLS, ODLD, OSLD, and ODLS contain 1170, 378, 401, and 825 scene pairs, respectively, totaling 2774. Figure~\ref{fig:examples} presents representative registered tile pairs for these subsets and imaging-geometry pair categories.

Although the four subsets contain similar numbers of tile pairs, their pixel spacings and per-tile ground footprints differ; tile-pair counts therefore do not directly represent actual ground-coverage areas.

Scene pairs are distributed primarily over low- and mid-latitude land areas, extending from approximately $55^\circ$ S to $66^\circ$ N and excluding the polar regions. The overall extent is set mainly by the sun-synchronous Sentinel-1 and NISAR subsets, whereas Capella and Umbra, with mid-inclination orbits, are confined to lower latitudes. Across subsets, the high-latitude limit is governed by land distribution, effective overlap, and sampling rather than by orbital geometry alone. Capella is concentrated in North America and mid-latitude Eurasia, Umbra spans multiple continents, and Sentinel-1 and NISAR exhibit relatively balanced spatial distributions under the combined urban and global random-sampling~strategy.

The subsets differ in temporal coverage. Capella spans December 2020 to February 2026, Umbra spans February 2024 to August 2025, Sentinel-1 spans January 2024 to January 2026, and NISAR spans June to October 2026. Capella, Umbra, and Sentinel-1 overlap during 2024--2025, enabling cross-platform comparisons within the same time window; note that each subset consists of intra-platform pairs rather than cross-sensor pairs. The long-term repeated observations of Capella also support same-location time-series analysis.

Taken together, the four subsets have comparable tile-pair counts and complementary platform, geographic, and temporal coverage.

\subsubsection{Land-Cover Composition}

\begin{figure*}[!tb]
  \centering
  \vspace{-0.8em}
  \includegraphics[width=0.88\textwidth]{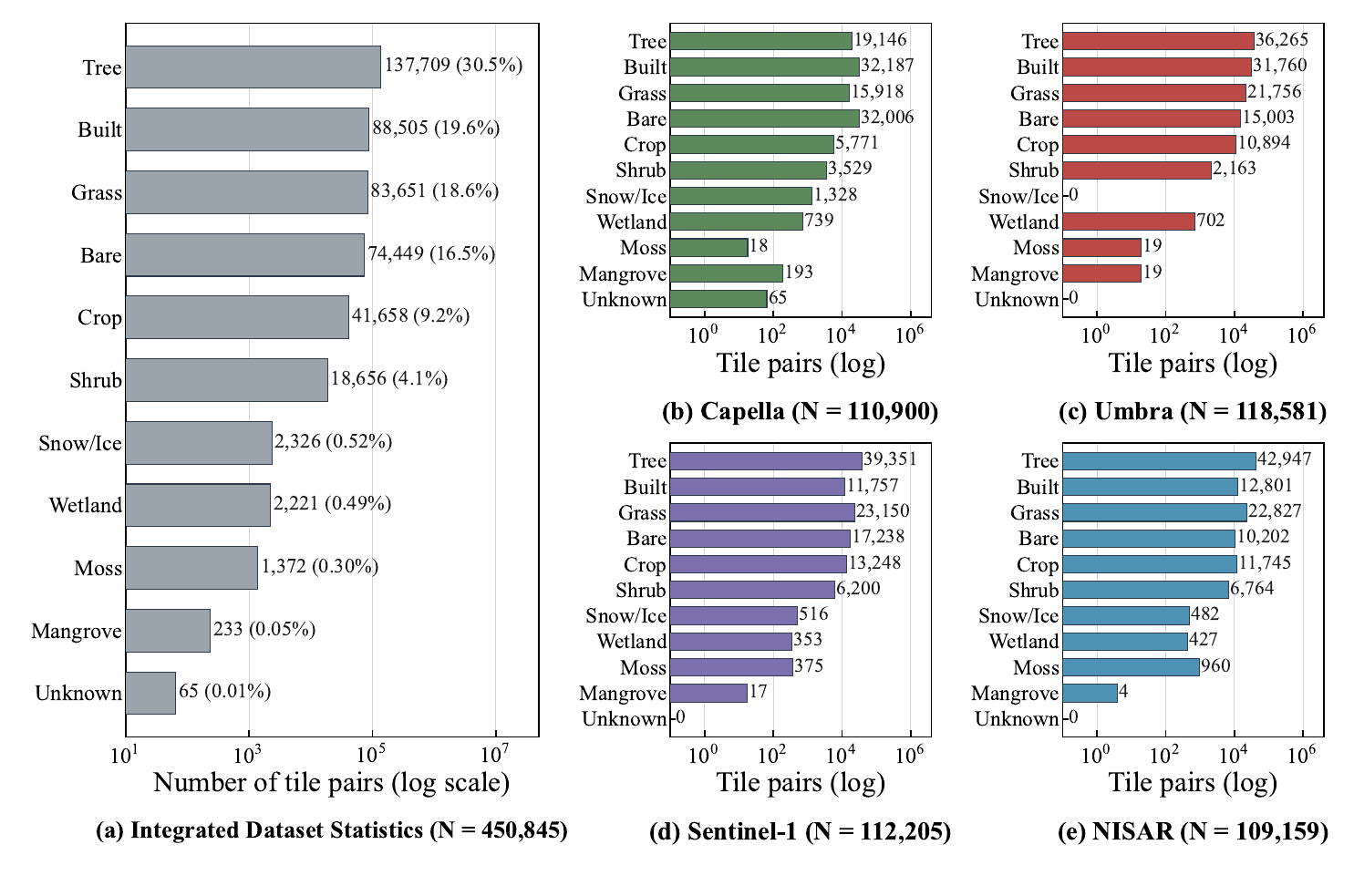}
  \caption{Dominant land-cover distributions in MOVR-SAR. (a) Class composition of all 450\,845 tile pairs, with percentages of the full dataset given in parentheses. (b)--(e) Class distributions in the Capella, Umbra, Sentinel-1, and NISAR subsets, respectively. All horizontal axes use logarithmic scales, and $N$ denotes the total number of tile pairs in each panel.}
  \label{fig:categorystats}
\end{figure*}

\label{sec:landcover}

Figure~\ref{fig:categorystats} summarizes the dominant land-cover classes across the full dataset and the four subsets. Tree cover, built-up areas, grassland, and bare or sparse vegetation jointly account for 85.2\% of all tile pairs. Their individual shares are 30.5\%, 19.6\%, 18.6\%, and 16.5\%, respectively. Cropland and shrubland account for 9.2\% and 4.1\%, while snow and ice, wetlands, moss and lichen, mangroves, and unknown areas each account for less than 1\%. Combining tree cover, grassland, and shrubland into a natural (non-cropland) vegetation group yields 53.2\%, and built-up areas plus bare or sparse vegetation (hard or sparsely vegetated surfaces) account for 36.1\%.

Class composition differs markedly among subsets. In Capella, built-up areas form the largest class at 29.0\%, essentially tied with bare or sparse vegetation at 28.9\% (within rounding). In Umbra, tree cover is the largest class at 30.6\%, followed by built-up areas at 26.8\%. Tree cover accounts for 35.1\% and 39.3\% of Sentinel-1 and NISAR, respectively, whereas built-up areas account for only 10.5\% and 11.7\%. Capella's tree-cover share is correspondingly the smallest of the four subsets. Thus, relative to the coarse-resolution subsets, the two high-resolution subsets have larger built-up shares, although their dominant classes differ (Capella: built-up; Umbra: tree cover), whereas the two coarser-resolution subsets are more strongly dominated by tree cover. Although Sentinel-1 and NISAR were sampled around urban areas, their much larger per-tile footprints (10--20 km) dilute the built-up signature, so few tiles receive a built-up dominant label; conversely, the sub-kilometre Capella and Umbra tiles more often fall entirely within built-up patches.

\begin{figure*}[!tb]
  \centering
  \includegraphics[width=0.92\textwidth]{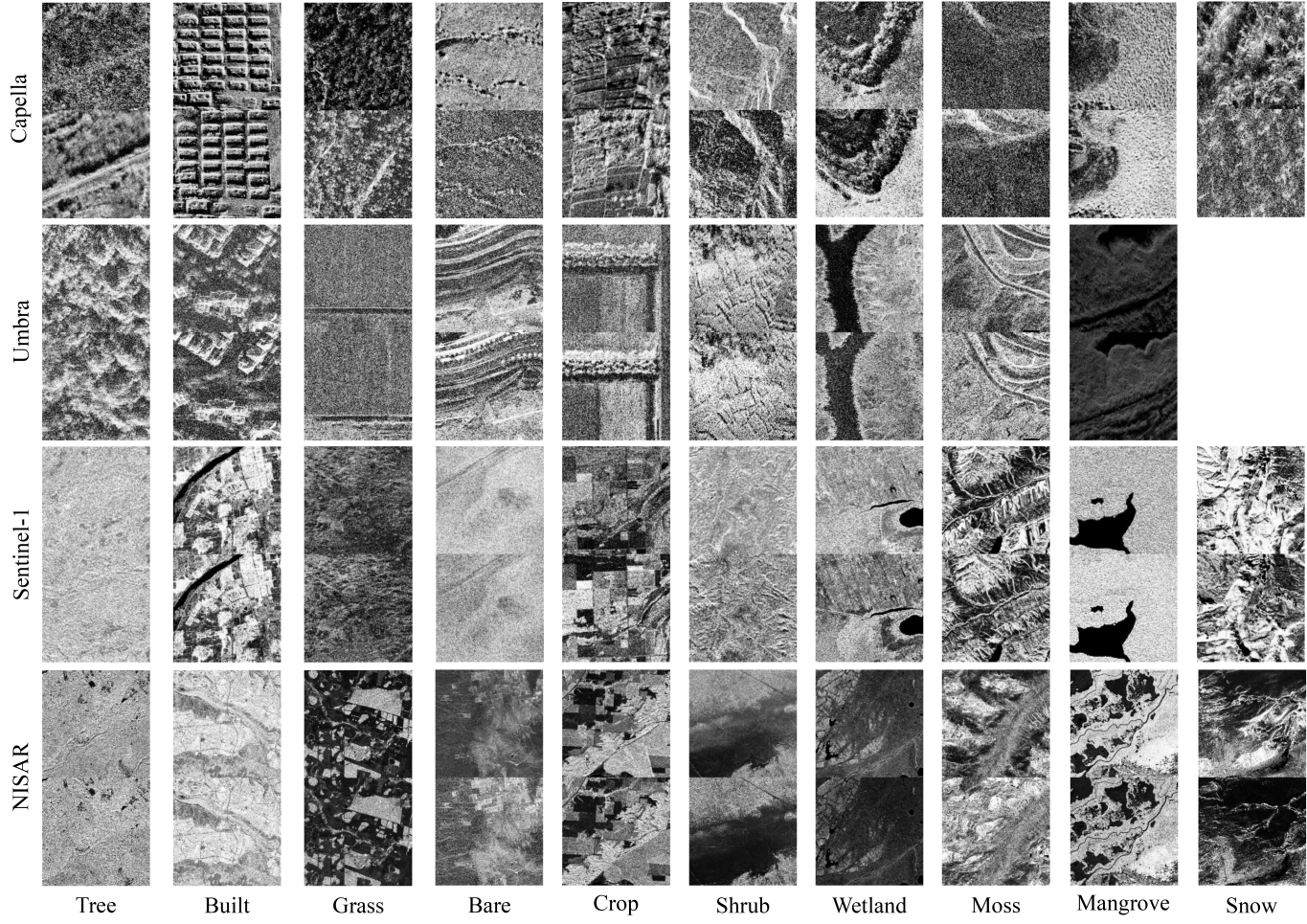}
  \caption{Representative SAR tile pairs for ten land-cover classes in MOVR-SAR. Rows correspond to Capella, Umbra, Sentinel-1, and NISAR, and columns correspond to the land-cover classes.}
  \label{fig:landcover_examples}
\end{figure*}

Overall, MOVR-SAR covers diverse surface types, but its class balance is subset-dependent: the high-resolution subsets contain larger built-up shares, whereas Sentinel-1 and NISAR are more strongly dominated by tree cover. These differences reflect the combined effects of sampling regions, underlying land-cover distributions, and tile ground footprints rather than sensor characteristics alone. Moreover, the reported percentages describe tile-pair counts by dominant class, not actual ground-area proportions. Consequently, cross-subset performance comparisons should account for land-cover composition and be complemented by class-stratified evaluation; sample sizes should also be reported, particularly for rare classes, to avoid unstable statistical conclusions. Figure~\ref{fig:landcover_examples} provides qualitative examples of image appearance across classes and platforms and is not used to infer class proportions.

%---------------------------------------------------------------------
\subsubsection{Viewing-Angle Separation Distribution}
\label{sec:angle}

The distributions of the viewing-angle separation $\Delta\theta$ by imaging-geometry pair category (panels~d and e) use the scene-level, nominal-incidence-angle approximation of the line-of-sight separation $\Omega$ defined in Eq.~\eqref{eq:dtheta}. Throughout this subsection, angles and their percentages are computed over scene pairs, whereas tile-pair counts are reported separately to indicate sample volume. Each row is normalized by the number of scene pairs in that category, allowing within-geometry angle distributions to be compared without confounding by group size. The same-direction OSLS and ODLD pairs all fall within $0^\circ$--$40^\circ$. In the $0^\circ$--$10^\circ$ bin, OSLS and ODLD account for 68.3\% and 49.4\% of Capella pairs, and 64.4\% and 63.9\% of Umbra pairs, respectively. Opposite-direction OSLD and ODLS pairs lie predominantly above $40^\circ$, with peaks concentrated between $60^\circ$ and $90^\circ$.

The angular supports of the two illumination groups are nearly disjoint. Among Capella and Umbra, the two subsets for which per-scene incidence angles are available so that $\Delta\theta$ can be computed (see Sec.~\ref{ssub:taxonomy}), all 959 same-direction scene pairs have $\Delta\theta<40^\circ$, whereas only 6 of the 668 opposite-direction pairs (approximately 0.9\%) fall below $40^\circ$. For Capella, the median angles of OSLS and ODLD are $4.9^\circ$ and $10.0^\circ$; for Umbra, they are $7.0^\circ$ and $7.1^\circ$. Among the opposite-direction categories, the median OSLD and ODLS angles are $73.5^\circ$ and $71.5^\circ$ in Capella, and $72.4^\circ$ and $74.2^\circ$ in Umbra. The binary illumination grouping therefore agrees closely with a $40^\circ$ threshold, although the small number of low-angle opposite-direction pairs shows that the two criteria are not strictly equivalent.

Taken together, Capella and Umbra provide substantial sample coverage for angle-aware evaluation, with 69\,232 same-direction and 41\,668 opposite-direction tile pairs in Capella, and 72\,391 and 46\,190 in Umbra, respectively. The nearly disjoint angular supports also justify $40^\circ$ as an approximate boundary between the two illumination regimes. However, viewing-angle separation is strongly coupled with imaging-geometry pair category, so between-category performance differences cannot be attributed to viewing angle alone. Angular effects should therefore be evaluated within each category or through joint stratification by imaging-geometry pair category and $\Delta\theta$.

%---------------------------------------------------------------------
\subsection{Dataset Characteristics}
\label{sec:characteristics}

The preceding statistics support four higher-level characteristics of MOVR-SAR.

(1) \textbf{Cross-platform, cross-resolution comparability.} Balanced tile-pair counts across platforms covering the X, C, and L bands at 0.25--20 m make cross-platform performance differences attributable to sensor and resolution rather than to sample size.

(2) \textbf{Spatiotemporal representativeness.} Global low- and mid-latitude coverage together with heterogeneous, partially overlapping time windows (including multi-year Capella observations) supports robustness and temporal-shift analysis without confounding geography with sensor.

(3) \textbf{Land-cover-aware evaluation.} Diverse surface types with subset-specific compositions allow matching difficulty to be assessed per land-cover class rather than only in aggregate.

(4) \textbf{Geometry-conditioned interpretability.} The four imaging-geometry pair categories and the viewing-angle separation jointly encode same- versus opposite-direction illumination, with a near-disjoint angular support, so that geometry effects can be attributed rather than confounded with viewing angle.

Together, these properties make MOVR-SAR a benchmark in which platform, resolution, land cover, and imaging geometry can be varied and attributed independently.

% =====================================================================
% ch5/ch5.tex
% Section 5: Experiments
% =====================================================================

\section{Experiments}\label{sec:experiments}
\sloppy

This section evaluates representative SAR image-matching methods on MOVR-SAR under a unified protocol across platforms, imaging-geometry pair categories, land-cover types, and viewing-angle separations, and measures the effect of fine-tuning. Here, image matching estimates correspondences for image registration, and matching accuracy based on homography residuals serves as a proxy for registration quality.

\subsection{Implementation Details}\label{sec:exp_setup}

\subsubsection{Evaluation Data}

All experiments use the MOVR-SAR dataset. To prevent information leakage, the split is performed at the scene-pair level: all tile pairs derived from the same scene pair are assigned to a single split, ensuring that no scene pair appears in more than one of the training, validation, and test sets. Owing to limited computational resources, training uses a subset of 100\,000 pairs, comprising 40\,000 each from Capella and Umbra and 10\,000 each from Sentinel-1 and NISAR. Within each platform, it approximately preserves the original distributions of imaging-geometry pair categories and land-cover types. This non-uniform allocation reflects differences in matching difficulty, assigning more samples to the challenging Capella and Umbra subsets and fewer to Sentinel-1 and NISAR. Despite the reduced size and non-uniform allocation, these 100\,000 pairs retain the geographic, land-cover, and imaging-geometry diversity of MOVR-SAR and remain representative of the full dataset. The complete validation set contains 400 pairs, evenly divided among the four platforms (100 each), and is used for model selection and hyperparameter tuning. The test sets are reserved exclusively for final evaluation and are not involved in training or model selection.

Two test sets support the different analyses. The overall, imaging-geometry, and land-cover experiments share a single set of 4000 pairs, balanced in imaging geometry: 250 pairs from each imaging-geometry pair category for Capella and Umbra, and 500 pairs from each of the two categories present in Sentinel-1 and NISAR, giving 1000 pairs per platform. This balanced composition supports per-category comparison without confounding by group size. The viewing-angle experiment instead uses a separate 4000-pair angle-test subset, evenly divided between Capella and Umbra (2000 each), the two platforms providing incidence angles. This subset is restricted to built-up tiles to hold land cover constant and make angle-dependent distortion directly observable. Both evaluation sets retain the same category assignments, enabling comparisons across categories and angular conditions on fixed samples.

All pairs in every split are subjected to the same random geometric-perturbation protocol: the relative rotation is at most $30^\circ$, the scale factor ranges from 1.2 to 1.3, and the translation is at most 5\% of the image side length. These operations do not change sample category assignments or the dataset taxonomy.

\subsubsection{Evaluation Metrics}

Correspondence quality is evaluated using the number of correct matches (NCM), inlier ratio (IR), root mean square error (RMSE), and success rate (SR), while registration accuracy is measured by the area under the curve (AUC). For each predicted correspondence, we compute a symmetric transfer error by projecting its endpoints with the ground-truth homography and its inverse. NCM counts valid correspondences with errors below 3 pixels, excluding those with endpoints outside the valid image content. IR is the ratio of NCM to all predicted correspondences. To keep correspondence counts comparable across methods, every method is restricted to at most 5000 predicted correspondences per image pair, keeping only the highest-confidence matches. RMSE is computed over the errors of these correct correspondences; if $\mathrm{NCM}<20$, RMSE is set to 20 pixels as a failure penalty. SR denotes the percentage of image pairs with $\mathrm{NCM}>20$. Higher NCM, IR, and SR indicate better matching, whereas lower RMSE indicates better geometric consistency.

For registration accuracy, a homography is estimated from the predicted matches for each image pair, and its mean corner error relative to the ground-truth homography is computed. The cumulative percentage of successfully registered pairs is evaluated at error thresholds of 5, 10, and 20 pixels, and AUC is calculated from this cumulative distribution. A higher AUC indicates more accurate and geometrically consistent registration. IR, SR, and AUC are reported as percentages, and RMSE is reported in pixels.

\subsubsection{Evaluated Methods}

The evaluated methods are grouped by how correspondences are generated and how densely they cover the image. Sparse methods establish correspondences by detecting and describing discrete keypoints; this category includes HIMO\cite{gao2026himo}, RIFT\cite{li2020rift}, LightGlue\cite{lindenberger2023lightglue}, MambaGlue\cite{ryoo2025mambaglue}, and MapGlue\cite{wu2026mapglue}. Semi-dense methods forgo explicit keypoint detection and establish selective correspondences through coarse-to-fine feature matching; this category includes ELoFTR\cite{wang2024eloftr}, GeoSAM\cite{zhang2026geosam}, XoFTR\cite{tuzcuouglu2024xoftr}, and FlexGlue\cite{lindenberger2026flexglue}. Dense methods estimate pixel-level correspondences, with RoMa v2\cite{edstedt2025romav2} included as the representative method. FlexGlue supports matching modes ranging from sparse to dense; because it starts from sparse seeds and produces a limited, selective set of correspondences rather than a full dense field, we classify it as semi-dense. RIFT and HIMO are evaluated with parameters tuned for this benchmark rather than their official defaults, which are not adapted to SAR--SAR matching under varying imaging geometry.

To distinguish the effects of matcher design from those of training data, the learning-based methods are evaluated with three weight sources: official weights, general-purpose cross-modal pretrained weights, and MOVR-SAR fine-tuned weights. The handcrafted methods (RIFT and HIMO) have no trained weights and are evaluated with their default parameter settings. The resulting 23 configurations comprise ten configurations in their original form (official weights or default parameters), six cross-modal configurations, and seven MOVR-SAR fine-tuned configurations. The cross-modal configurations use the LightGlue, ELoFTR, XoFTR, and RoMa v2 weights from MINIMA\cite{ren2025minima}, together with the ELoFTR and RoMa v2 weights from MatchAnything\cite{he2025matchanything}. Because MINIMA and MatchAnything are general-purpose cross-modal training frameworks rather than standalone matching methods, their configurations are categorized according to the underlying backbone. Fine-tuning is performed from the official weights on four NVIDIA RTX 4090 GPUs for 10 epochs; the checkpoint with the best validation-set performance is selected as the final weights. All fine-tuning experiments use the same training set and a uniform input tile size equal to the released tile size. Numerical overall results are available for 22 configurations; the official GeoSAM configuration is retained in the results table with blank cells.

\subsection{Overall Performance}\label{sec:exp_dataset}

\begin{table*}[t]
\centering
\tabcaption{Overall correspondence quality on AUC}
\label{tab:auc_dataset}
\footnotesize
\setlength{\tabcolsep}{1.8pt}
\renewcommand{\arraystretch}{1.06}
\newcommand{\ftcell}[1]{\textcolor{blue}{#1}}
\newcommand{\bestcell}[1]{\textcolor{blue}{\textbf{#1}}}
\begin{threeparttable}
\begin{tabular}{l l
  @{\hspace{1.5pt}} c @{\hspace{1.0pt}} c @{\hspace{1.0pt}} c
  @{\hspace{2.0pt}} c @{\hspace{1.0pt}} c @{\hspace{1.0pt}} c
  @{\hspace{2.0pt}} c @{\hspace{1.0pt}} c @{\hspace{1.0pt}} c
  @{\hspace{2.0pt}} c @{\hspace{1.0pt}} c @{\hspace{1.0pt}} c}
\toprule
\multirow{2}{*}{Type} & \multirow{2}{*}{Method} & \multicolumn{3}{c}{Capella} & \multicolumn{3}{c}{Umbra} & \multicolumn{3}{c}{Sentinel-1} & \multicolumn{3}{c}{NISAR} \\
\cmidrule(lr){3-5}\cmidrule(lr){6-8}\cmidrule(lr){9-11}\cmidrule(lr){12-14}
 & & AUC@5 & AUC@10 & AUC@20 & AUC@5 & AUC@10 & AUC@20 & AUC@5 & AUC@10 & AUC@20 & AUC@5 & AUC@10 & AUC@20 \\
\midrule
\multirow{8}{*}{Sparse} & RIFT (TIP 20) & 0.12 & 0.29 & 0.46 & 0.11 & 0.32 & 0.73 & 10.08 & 15.02 & 19.80 & 50.92 & 62.49 & 70.48 \\
 & HIMO (TPAMI 26) & 0.10 & 0.15 & 0.21 & 0.00 & 0.00 & 0.00 & 3.62 & 7.42 & 10.92 & 23.81 & 36.70 & 46.19 \\
 & LightGlue (ICCV 23) & 0.40 & 0.96 & 1.86 & 0.22 & 0.53 & 1.13 & 6.92 & 13.18 & 20.03 & 40.39 & 57.78 & 69.34 \\
 & MambaGlue (ICRA 25) & 0.75 & 2.20 & 4.35 & 0.41 & 1.39 & 3.54 & 14.67 & 29.08 & 43.13 & 49.08 & 68.01 & 79.65 \\
 & MapGlue (ISPRS 26) & 0.91 & 2.60 & 5.51 & 0.55 & 2.54 & 6.73 & 19.73 & 36.73 & 52.14 & 53.82 & 72.56 & 83.66 \\
 & $\mathrm{MINIMA}_{\mathrm{LightGlue}}$ & 0.31 & 0.88 & 1.85 & 0.18 & 0.40 & 0.95 & 5.91 & 11.90 & 17.95 & 40.30 & 57.48 & 68.61 \\
 & \ftcell{$\mathrm{MOVR\text{-}SAR}_{\mathrm{LightGlue}}$} & \ftcell{0.89} & \ftcell{6.22} & \ftcell{21.67} & \ftcell{0.19} & \ftcell{3.50} & \ftcell{14.48} & \ftcell{9.07} & \ftcell{22.49} & \ftcell{40.18} & \ftcell{43.14} & \ftcell{63.28} & \ftcell{77.58} \\
 & \ftcell{$\mathrm{MOVR\text{-}SAR}_{\mathrm{MambaGlue}}$} & \ftcell{20.71} & \ftcell{50.67} & \ftcell{73.71} & \ftcell{22.02} & \ftcell{50.54} & \ftcell{73.67} & \ftcell{32.55} & \ftcell{61.33} & \ftcell{80.33} & \ftcell{53.30} & \ftcell{75.05} & \ftcell{87.36} \\
\midrule
\multirow{11}{*}{\makecell[l]{Semi-\\dense}} & ELoFTR (CVPR 24) & 0.64 & 1.35 & 2.35 & 0.26 & 1.03 & 2.50 & 8.93 & 14.49 & 20.71 & 45.39 & 56.72 & 64.86 \\
 & XoFTR (CVPR 24) & 0.29 & 0.42 & 0.79 & 0.08 & 0.27 & 0.68 & 6.67 & 10.80 & 15.03 & 40.09 & 51.05 & 59.32 \\
 & GeoSAM (ISPRS 26) & 0.43 & 0.79 & 2.47 & 0.24 & 0.58 & 1.58 & 6.13 & 13.07 & 29.73 & 9.85 & 18.82 & 30.26 \\
 & FlexGlue (arXiv 26) & 3.06 & 5.74 & 9.24 & 1.79 & 5.14 & 9.80 & 42.69 & 55.92 & 65.72 & 77.68 & 86.23 & 90.92 \\
 & $\mathrm{MINIMA}_{\mathrm{ELoFTR}}$ & 0.12 & 0.21 & 0.50 & 0.14 & 0.38 & 1.03 & 2.86 & 5.79 & 9.32 & 30.05 & 43.22 & 53.62 \\
 & $\mathrm{MINIMA}_{\mathrm{XoFTR}}$ & 0.07 & 0.08 & 0.15 & 0.00 & 0.00 & 0.00 & 1.89 & 3.57 & 5.36 & 28.12 & 37.81 & 45.37 \\
 & $\mathrm{MatchAnything}_{\mathrm{ELoFTR}}$ & 1.38 & 3.02 & 5.60 & 0.76 & 2.25 & 5.19 & 26.32 & 38.75 & 49.22 & 70.09 & 81.09 & 87.66 \\
 & \ftcell{$\mathrm{MOVR\text{-}SAR}_{\mathrm{ELoFTR}}$} & \ftcell{49.36} & \ftcell{68.86} & \ftcell{83.69} & \ftcell{46.03} & \ftcell{66.32} & \ftcell{82.12} & \ftcell{59.91} & \ftcell{76.86} & \ftcell{87.64} & \ftcell{80.64} & \ftcell{89.57} & \ftcell{94.33} \\
 & \ftcell{$\mathrm{MOVR\text{-}SAR}_{\mathrm{XoFTR}}$} & \ftcell{65.32} & \ftcell{80.56} & \ftcell{88.88} & \ftcell{65.12} & \ftcell{80.08} & \ftcell{88.38} & \ftcell{66.11} & \ftcell{81.04} & \ftcell{89.58} & \ftcell{65.47} & \ftcell{80.64} & \ftcell{89.27} \\
 & \ftcell{$\mathrm{MOVR\text{-}SAR}_{\mathrm{GeoSAM}}$} & \ftcell{41.32} & \ftcell{59.12} & \ftcell{73.68} & \ftcell{38.44} & \ftcell{56.26} & \ftcell{71.09} & \ftcell{41.46} & \ftcell{60.23} & \ftcell{74.92} & \ftcell{45.03} & \ftcell{62.72} & \ftcell{75.69} \\
 & \ftcell{$\mathrm{MOVR\text{-}SAR}_{\mathrm{FlexGlue}}$} & \bestcell{89.79} & \bestcell{94.76} & \bestcell{97.38} & \bestcell{89.03} & \bestcell{94.37} & \bestcell{97.19} & \bestcell{93.41} & \bestcell{96.69} & \bestcell{98.34} & \bestcell{94.77} & \bestcell{97.39} & \bestcell{98.69} \\
\midrule
\multirow{4}{*}{Dense} & RoMa v2 (ECCV 26) & 2.77 & 5.57 & 9.64 & 1.79 & 5.18 & 10.21 & 43.20 & 55.53 & 65.54 & 81.75 & 87.97 & 91.60 \\
 & $\mathrm{MINIMA}_{\mathrm{RoMa\;v2}}$ & 0.00 & 0.00 & 0.06 & 0.00 & 0.00 & 0.00 & 0.09 & 0.22 & 0.49 & 14.51 & 20.11 & 25.83 \\
 & $\mathrm{MatchAnything}_{\mathrm{RoMa\;v2}}$ & 1.36 & 2.76 & 4.82 & 0.55 & 1.43 & 3.52 & 23.80 & 32.00 & 39.11 & 60.82 & 68.18 & 73.25 \\
 & \ftcell{$\mathrm{MOVR\text{-}SAR}_{\mathrm{RoMa\;v2}}$} & \ftcell{74.82} & \ftcell{86.88} & \ftcell{93.42} & \ftcell{73.91} & \ftcell{86.66} & \ftcell{93.33} & \ftcell{82.64} & \ftcell{91.32} & \ftcell{95.66} & \ftcell{89.28} & \ftcell{94.62} & \ftcell{97.31} \\
\bottomrule
\end{tabular}
\begin{tablenotes}[flushleft]
\footnotesize
\item \textcolor{blue}{Blue} denotes MOVR-SAR fine-tuned configurations, and \textcolor{blue}{\textbf{bold blue}} denotes the best result in each column.
\end{tablenotes}
\end{threeparttable}
\end{table*}

Tables~\ref{tab:auc_dataset} and~\ref{tab:nris_dataset} report registration AUC and correspondence quality (NCM, RMSE, IR, SR) for the configurations with available results. The clearest pattern is the strong platform dependence of the original configurations: on the high-resolution X-band platforms (Capella and Umbra) essentially all configurations with original settings collapse, with NCM of only a few correspondences and RMSE close to the failure penalty, so that usable correspondences are rarely established. Performance is substantially higher on the coarser-resolution Sentinel-1 and NISAR, so these platforms alone would overestimate general SAR-matching capability. NISAR also performs better than Sentinel-1, which is largely attributable to their radar bands: NISAR operates at L-band and Sentinel-1 at C-band, and the longer L-band wavelength penetrates the vegetation canopy and interacts more with the ground and trunks, reducing vegetation-induced image variation so that L-band backscatter is more stable for matching over vegetated terrain, whereas the shorter C-band primarily senses the canopy. Because the two platforms also differ in resolution and acquisition conditions, this is a conditional association rather than a strict causal proof.

The original configurations also reveal broad differences among matching paradigms: on Sentinel-1 and NISAR, dense and detector-free matchers such as RoMa v2 and FlexGlue perform comparatively well, whereas sparse keypoint matchers such as LightGlue and HIMO lag substantially, and RIFT is competitive on NISAR but collapses on both X-band platforms. General-purpose cross-modal pretraining does not resolve this limitation, as MINIMA variants are often weaker than the corresponding official weights and MatchAnything variants still fail on~X-band.

Fine-tuning on MOVR-SAR produces the dominant improvement, with the largest gains on the X-band platforms where official and cross-modal weights fail; fine-tuned FlexGlue and RoMa v2 exceed 99\% SR on all four platforms with RMSE below 1 pixel. The resulting ranking is clear: FlexGlue and RoMa v2 are strongest overall, XoFTR and ELoFTR follow, GeoSAM and MambaGlue are intermediate, and fine-tuned LightGlue remains weak. These gains should be distinguished from absolute performance, since configurations already strong on Sentinel-1 and NISAR improve comparatively little, so improvement magnitude alone does not establish model superiority.

Fine-tuning also reduces within-method variation across platforms, though this should be read together with the non-uniform training allocation, which provides four times more Capella and Umbra samples than Sentinel-1 and NISAR samples. Overall, platform and weight source are the main performance factors, with fine-tuned FlexGlue and RoMa v2 providing the strongest accuracy and correspondence quality across the benchmark.

\begin{table*}[t]
\centering
\tabcaption{Overall correspondence quality on NCM, RMSE, IR, SR}
\label{tab:nris_dataset}
\footnotesize
\setlength{\tabcolsep}{2.0pt}
\renewcommand{\arraystretch}{1.06}
\newcommand{\ftcell}[1]{\textcolor{blue}{#1}}
\newcommand{\bestcell}[1]{\textcolor{blue}{\textbf{#1}}}
\newcommand{\bestplain}[1]{\textbf{#1}}
\begin{threeparttable}
\begin{tabular*}{\textwidth}{@{} l @{\extracolsep{\fill}} *{16}{c} @{}}
\toprule
\multirow{2}{*}{Method} & \multicolumn{4}{c}{Capella} & \multicolumn{4}{c}{Umbra} & \multicolumn{4}{c}{Sentinel-1} & \multicolumn{4}{c}{NISAR} \\
\cmidrule(lr){2-5}\cmidrule(lr){6-9}\cmidrule(lr){10-13}\cmidrule(lr){14-17}
 & NCM & RMSE & IR & SR & NCM & RMSE & IR & SR & NCM & RMSE & IR & SR & NCM & RMSE & IR & SR \\
\midrule
RIFT (TIP 20) & 0.20 & 19.98 & 0.05 & 0.10 & 0.24 & 20.00 & 0.06 & 0.00 & 5.55 & 18.38 & 1.23 & 8.20 & 40.63 & 9.21 & 9.74 & 57.90 \\
HIMO (TPAMI 26) & 0.18 & 19.98 & 0.20 & 0.10 & 0.02 & 20.00 & 0.08 & 0.00 & 7.17 & 17.96 & 8.67 & 10.70 & 48.91 & 10.38 & 39.49 & 52.50 \\
LightGlue (ICCV 23) & 2.82 & 19.31 & 1.96 & 3.60 & 1.49 & 19.64 & 1.96 & 2.00 & 27.31 & 15.03 & 19.83 & 26.40 & 172.10 & 4.85 & 61.04 & 82.10 \\
MambaGlue (ICRA 25) & 4.08 & 18.59 & 4.34 & 7.30 & 4.50 & 18.36 & 5.08 & 8.50 & 48.36 & 8.35 & 44.23 & 63.50 & 139.75 & 2.71 & 81.84 & 93.30 \\
MapGlue (ISPRS 26) & 34.01 & 17.00 & 3.56 & 16.70 & 40.15 & 15.89 & 4.32 & 22.80 & 315.50 & 5.48 & 29.22 & 80.30 & 652.22 & 2.30 & 56.12 & 96.90 \\
$\mathrm{MINIMA}_{\mathrm{LightGlue}}$ & 2.54 & 19.42 & 1.63 & 3.00 & 1.45 & 19.64 & 1.65 & 1.80 & 26.98 & 15.00 & 17.59 & 27.10 & 172.81 & 5.03 & 60.90 & 81.20 \\
\ftcell{$\mathrm{MOVR\text{-}SAR}_{\mathrm{LightGlue}}$} & \ftcell{21.73} & \ftcell{13.55} & \ftcell{9.89} & \ftcell{34.90} & \ftcell{12.70} & \ftcell{15.95} & \ftcell{7.95} & \ftcell{21.10} & \ftcell{46.11} & \ftcell{9.91} & \ftcell{22.77} & \ftcell{54.10} & \ftcell{189.93} & \ftcell{3.36} & \ftcell{50.40} & \ftcell{90.60} \\
\ftcell{$\mathrm{MOVR\text{-}SAR}_{\mathrm{MambaGlue}}$} & \ftcell{91.23} & \ftcell{2.41} & \ftcell{69.56} & \ftcell{97.90} & \ftcell{92.48} & \ftcell{2.25} & \ftcell{70.81} & \ftcell{98.70} & \ftcell{166.59} & \ftcell{1.95} & \ftcell{80.11} & \ftcell{99.80} & \ftcell{314.06} & \ftcell{1.80} & \ftcell{88.54} & \ftcell{99.40} \\
\midrule
ELoFTR (CVPR 24) & 7.42 & 19.13 & 1.34 & 4.60 & 6.24 & 18.85 & 1.39 & 6.30 & 81.55 & 14.21 & 10.89 & 30.60 & 572.10 & 5.76 & 43.98 & 76.30 \\
XoFTR (CVPR 24) & 4.20 & 19.37 & 0.84 & 3.10 & 2.91 & 19.57 & 0.77 & 2.30 & 59.89 & 14.69 & 8.25 & 27.90 & 481.41 & 6.03 & 35.89 & 74.80 \\
GeoSAM (ISPRS 26) & 6.42 & 19.21 & 1.53 & 4.15 & 4.18 & 19.36 & 1.27 & 2.85 & 74.63 & 14.42 & 8.67 & 31.28 & 197.54 & 9.63 & 24.18 & 58.74 \\
FlexGlue (arXiv 26) & 389.36 & 15.61 & 8.95 & 24.10 & 396.01 & 14.97 & 9.17 & 27.70 & 3206.27 & 4.26 & 66.75 & 84.30 & 4547.32 & 1.27 & 91.65 & 97.30 \\
$\mathrm{MINIMA}_{\mathrm{ELoFTR}}$ & 2.08 & 19.69 & 1.09 & 1.70 & 3.78 & 19.28 & 1.70 & 4.00 & 36.77 & 16.36 & 8.83 & 19.80 & 419.89 & 6.78 & 39.87 & 71.90 \\
$\mathrm{MINIMA}_{\mathrm{XoFTR}}$ & 0.57 & 19.98 & 1.35 & 0.10 & 0.11 & 19.98 & 1.11 & 0.10 & 5.10 & 19.24 & 13.02 & 4.10 & 180.09 & 10.66 & 45.83 & 49.50 \\
$\mathrm{MatchAnything}_{\mathrm{ELoFTR}}$ & 26.44 & 18.01 & 3.91 & 10.90 & 20.09 & 17.32 & 3.77 & 14.40 & 331.07 & 7.90 & 35.53 & 65.50 & 1296.82 & 2.04 & 73.87 & 95.40 \\
\ftcell{$\mathrm{MOVR\text{-}SAR}_{\mathrm{ELoFTR}}$} & \ftcell{1723.13} & \ftcell{1.35} & \ftcell{80.93} & \ftcell{99.70} & \ftcell{1653.01} & \ftcell{1.48} & \ftcell{78.39} & \ftcell{99.30} & \ftcell{1764.36} & \ftcell{1.32} & \ftcell{84.56} & \ftcell{99.30} & \ftcell{1928.02} & \ftcell{0.97} & \ftcell{92.30} & \ftcell{99.50} \\
\ftcell{$\mathrm{MOVR\text{-}SAR}_{\mathrm{XoFTR}}$} & \ftcell{3269.05} & \ftcell{1.65} & \ftcell{93.73} & \ftcell{97.50} & \ftcell{3259.17} & \ftcell{1.64} & \ftcell{94.11} & \ftcell{97.70} & \ftcell{3192.18} & \ftcell{1.56} & \ftcell{92.79} & \ftcell{98.00} & \ftcell{3132.49} & \ftcell{1.61} & \ftcell{90.90} & \ftcell{97.70} \\
\ftcell{$\mathrm{MOVR\text{-}SAR}_{\mathrm{GeoSAM}}$} & \ftcell{2317.42} & \ftcell{2.83} & \ftcell{70.97} & \ftcell{93.20} & \ftcell{2205.53} & \ftcell{2.91} & \ftcell{67.92} & \ftcell{92.70} & \ftcell{2267.18} & \ftcell{2.58} & \ftcell{69.63} & \ftcell{94.50} & \ftcell{2247.19} & \ftcell{2.69} & \ftcell{66.99} & \ftcell{93.60} \\
\ftcell{$\mathrm{MOVR\text{-}SAR}_{\mathrm{FlexGlue}}$} & \bestcell{4899.92} & \bestcell{0.40} & \bestcell{98.00} & \bestcell{99.90} & \bestcell{4875.76} & \bestcell{0.41} & \bestcell{97.52} & \bestcell{100.00} & \ftcell{4926.09} & \bestcell{0.28} & \ftcell{98.52} & \bestcell{100.00} & \ftcell{4898.54} & \bestcell{0.24} & \ftcell{97.97} & \bestcell{100.00} \\
\midrule
RoMa v2 (ECCV 26) & 423.01 & 14.55 & 8.46 & 30.00 & 455.99 & 13.68 & 9.12 & 34.90 & 3213.02 & 3.08 & 64.26 & 90.40 & 4532.12 & 0.98 & 90.64 & 97.90 \\
$\mathrm{MINIMA}_{\mathrm{RoMa\;v2}}$ & 3.29 & 19.30 & 0.07 & 3.50 & 8.36 & 18.33 & 0.17 & 9.20 & 27.15 & 17.12 & 0.54 & 15.50 & 955.43 & 7.85 & 19.11 & 65.50 \\
$\mathrm{MatchAnything}_{\mathrm{RoMa\;v2}}$ & 201.69 & 15.31 & 4.03 & 25.50 & 171.27 & 14.15 & 3.43 & 32.40 & 1664.84 & 7.14 & 33.30 & 69.50 & 3268.20 & 3.08 & 65.36 & 88.80 \\
\ftcell{$\mathrm{MOVR\text{-}SAR}_{\mathrm{RoMa\;v2}}$} & \ftcell{4825.81} & \ftcell{0.91} & \ftcell{96.52} & \ftcell{99.70} & \ftcell{4825.72} & \ftcell{0.93} & \ftcell{96.51} & \ftcell{99.80} & \bestcell{4972.05} & \ftcell{0.67} & \bestcell{99.44} & \bestcell{100.00} & \bestcell{4975.46} & \ftcell{0.47} & \bestcell{99.51} & \bestcell{100.00} \\
\bottomrule
\end{tabular*}
\begin{tablenotes}[flushleft]
\footnotesize
\item Blank cells indicate configurations without experimental results. Values are percentages. Subscripts identify the weight sources: MOVR-SAR denotes fine-tuning on MOVR-SAR, whereas MINIMA and MatchAnything denote the corresponding cross-modal pretrained weights. \textcolor{blue}{Blue} denotes MOVR-SAR fine-tuned configurations, and \textcolor{blue}{\textbf{bold blue}} denotes the best result in each column.
\end{tablenotes}
\end{threeparttable}
\end{table*}

\subsection{Imaging-Geometry Performance}\label{sec:exp_oxlx}

\begin{figure*}[!tb]
  \centering
  \includegraphics[width=\textwidth]{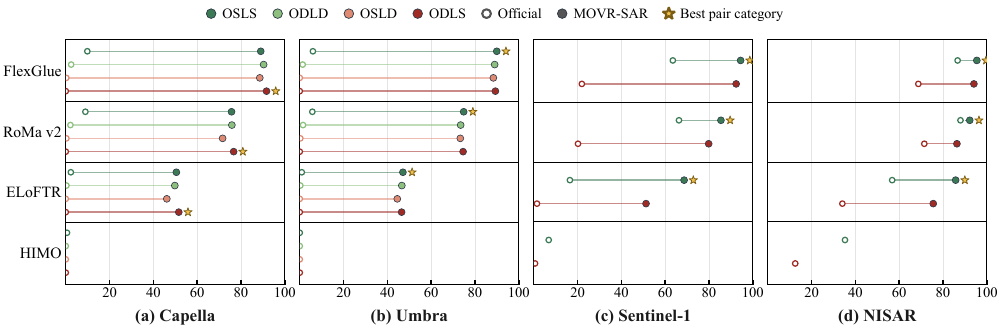}
  \caption{$\mathrm{AUC}\text{@}5$ by imaging-geometry pair category on the four platforms (a--d: Capella, Umbra, Sentinel-1, NISAR); categories run top to bottom as OSLS, ODLD, OSLD, and ODLS (Sentinel-1 and NISAR contain only OSLS and ODLS). Open and filled markers denote original and MOVR-SAR fine-tuned forms, joined by lines showing the fine-tuning gain; asterisks mark the best fine-tuned category. HIMO is a non-learning method and is evaluated with fixed, recommended parameters (open markers only); $N$ gives test pairs per panel.}
  \label{fig:auc_oxlx}
\end{figure*}

This section isolates performance differences among the four imaging-geometry pair categories. We compare ELoFTR, RoMa v2, and FlexGlue in their official and MOVR-SAR fine-tuned forms, together with HIMO under its recommended parameter setting, yielding seven configurations. All configurations use the same 4000 category-balanced test pairs described in the implementation details, with the four categories for Capella and Umbra and only OSLS and ODLS for the right-looking Sentinel-1 and NISAR. Figure~\ref{fig:auc_oxlx} reports $\mathrm{AUC}\text{@}5$ for each group. Although the four categories form two illumination-direction regimes (OSLS and ODLD: same direction; OSLD and ODLS: opposite direction), the four-category resolution is retained because regime-level averages can mask category-specific~behavior.

With the original settings, the dominant geometry effect is a consistent OSLS advantage. RoMa v2 achieves 8.96\% on Capella OSLS but 0\% on ODLS, and official FlexGlue similarly drops from 9.85\% to 0\%. Averaging ELoFTR, RoMa v2, and FlexGlue gives $\mathrm{AUC}\text{@}5$ values of 7.04\%, 1.63\%, 0.18\%, and 0\% for Capella OSLS, ODLD, OSLD, and ODLS, respectively, and 4.23\%, 1.03\%, 0.09\%, and 0\% for the corresponding Umbra categories. HIMO shows the same tendency on NISAR, where OSLS exceeds ODLS by 22.71 percentage points. Except for HIMO on Umbra, where all four results are 0\%, every method in its original form performs best on OSLS on the two X-band platforms. Performance with the original settings is therefore highest for same-orbit, same-side acquisitions and deteriorates when either orbit direction or look side changes.

After fine-tuning, the main category-level change is a substantial reduction in these geometry gaps. RoMa v2 improves from 0\% on ODLS to 76.66\% on Capella and 74.58\% on Umbra. Averaging the three paired methods yields 71.74\%, 72.00\%, 68.79\%, and 73.31\% for Capella OSLS, ODLD, OSLD, and ODLS, respectively, with only 4.52 percentage points between the highest and lowest values; the corresponding range on Umbra is 1.88 points. FlexGlue also becomes consistent across categories, ranging from 88.60\% to 91.67\% on Capella and from 88.41\% to 89.95\% on Umbra.

The category ranking after fine-tuning clarifies the origin of the original-setting bias. On Capella, the three-method average is highest on ODLS (73.31\%), reversing the original OSLS preference. Thus, the OSLS advantage is not an invariant measure of geometric difficulty but is influenced by training-data coverage: cross-orbit and cross-side samples can achieve high accuracy once all categories are represented during training. This reversal is nevertheless platform-specific: it is pronounced on Capella, absent on Umbra, and not observed on Sentinel-1 or NISAR. It should therefore be interpreted as an empirical category pattern, not a universal ranking.

A residual orbit-direction effect remains on Sentinel-1 and NISAR, where OSLS and ODLS share the same look side and thus isolate orbit-direction changes more directly. Fine-tuned ELoFTR and RoMa v2 still favor OSLS, with average gaps of approximately 14 and 6 percentage points across the two platforms, respectively; the corresponding FlexGlue gaps are only 2.04 and 1.33 points, compared with official-weight gaps of 15.0--46.1 points. The geometry-specific conclusion is therefore twofold: fine-tuning nearly eliminates category differences on the X-band platforms, but a smaller same-orbit advantage persists on Sentinel-1 and NISAR. Because these residual gaps may also involve scene content, overlap, and viewing-angle separation, the angular contribution is examined separately in the next subsection.

\subsection{Land-Cover Performance}\label{sec:exp_landcover}

\begin{figure*}[!tb]
  \centering
  \includegraphics[width=\textwidth]{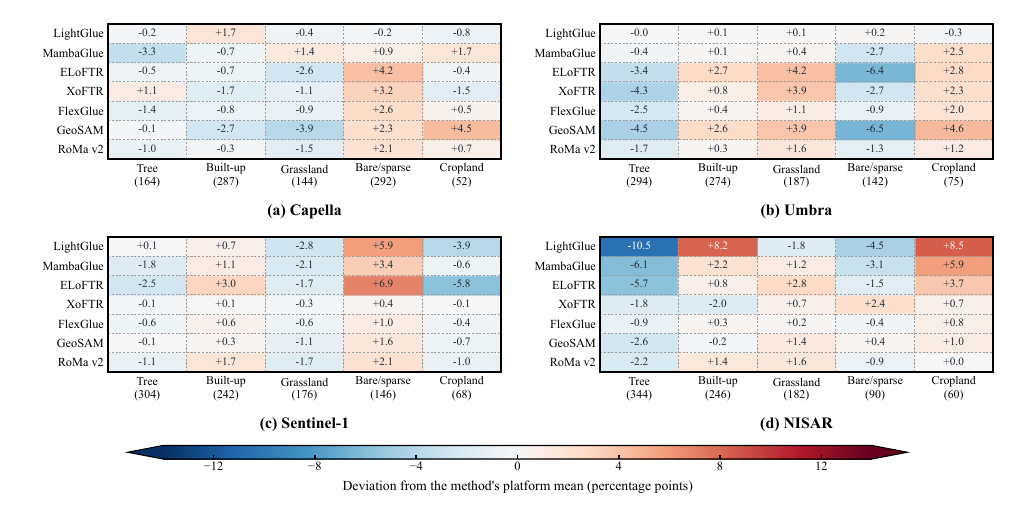}
  \caption{Land-cover-stratified $\mathrm{AUC}\text{@}5$ deviations for seven fine-tuned methods on the four platforms (a--d: Capella, Umbra, Sentinel-1, NISAR). Parenthesized values beneath the class labels are test-pair counts. Cell values and colors give each method's deviation from its five-class mean (percentage points); red and blue denote positive and negative deviations.}
  \label{fig:auc_landcover}
\end{figure*}

To assess the influence of land cover on matching performance, we evaluate MOVR-SAR fine-tuned models of LightGlue, MambaGlue, ELoFTR, XoFTR, FlexGlue, GeoSAM, and RoMa v2. The overall test set is grouped by land-cover label, using the same homography-error and AUC protocol as in the other experiments. Five classes are considered: \texttt{tree} (tree cover), \texttt{built} (built-up areas), \texttt{grass} (grassland), \texttt{bare} (bare or sparse vegetation), and \texttt{crop} (cropland). Each platform--class combination contains 52--344 pairs.

Figure~\ref{fig:auc_landcover} reveals two distinct patterns: differences in robustness among methods and platform-dependent class biases. FlexGlue and RoMa v2 are consistently stable across land-cover types, with within-platform class ranges of 1.65--4.43 and 3.34--3.87 percentage points, respectively. By contrast, LightGlue varies by 18.95 points across classes on NISAR, indicating substantially greater sensitivity. Stability across land cover is therefore complementary to, rather than a substitute for, absolute matching accuracy.

Averaged across the seven methods, \texttt{tree} is below the five-class mean on every platform, with the largest negative deviation on NISAR ($-4.25$ percentage points). However, the highest-performing class changes across platforms: \texttt{bare} on Capella and Sentinel-1 ($+2.15$ and $+3.06$ points), \texttt{grass} on Umbra ($+2.17$ points), and \texttt{crop} on NISAR ($+2.96$ points). Land-cover effects are thus platform- and method-dependent rather than following a universal difficulty ranking. Because class labels correlate with scene composition, image overlap, and imaging geometry, and sample sizes vary across classes, these deviations should be interpreted as conditional associations rather than isolated causal effects of land cover.

\subsection{Viewing-Angle Performance}\label{sec:exp_angle}

\begin{figure*}[!tb]
  \centering
  \includegraphics[width=\textwidth]{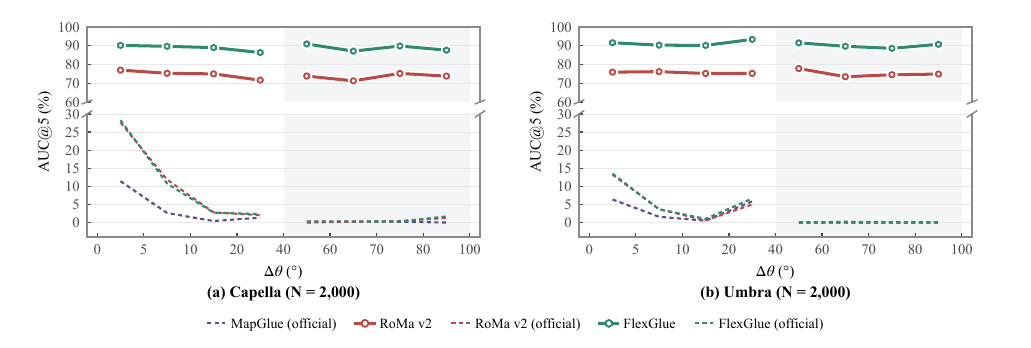}
  \caption{$\mathrm{AUC}\text{@}5$ versus viewing-angle separation on the angle-test subset for Capella (a) and Umbra (b). Solid and dashed lines denote MOVR-SAR fine-tuned and official weights, respectively (MapGlue uses official weights only); shading marks the opposite-direction regime ($\Delta\theta\geq40^\circ$). The vertical axis is broken between 30\% and 60\%, which contains no data.}
  \label{fig:auc_angle}
\end{figure*}

The angular analysis evaluates official and fine-tuned weights of RoMa v2 and FlexGlue, complemented by MapGlue with official weights, for a total of five configurations. All configurations use the same 4000 angle-test pairs, comprising 2000 each from Capella and Umbra. All selected pairs belong to the built-up (\texttt{built}) land-cover class; restricting the analysis to this single class keeps the other factors uniform, and built-up areas exhibit pronounced shadow and layover in every region, which makes the angle-dependent distortion directly observable. Pairs are divided into eight $\Delta\theta$ intervals, and $\mathrm{AUC}\text{@}5$ is computed separately for each platform. Only groups containing at least five pairs are plotted. In Figure~\ref{fig:auc_angle}, the shaded background denotes the opposite-direction illumination regime with $\Delta\theta\geq40^\circ$, an angular proxy for the categorical opposite-direction regime that is nearly, but not exactly, equivalent.

The official weights exhibit two clear patterns. First, accuracy is generally low, and in the opposite-direction regime ($\Delta\theta\geq40^\circ$) it falls close to zero for all three methods on both platforms, so almost no usable matches are produced once the two viewing directions diverge. Second, within the same-direction regime ($\Delta\theta<40^\circ$), matching becomes progressively harder as the angle separation grows: accuracy decreases monotonically with $\Delta\theta$ for every method and platform, and the drop is already substantial at small-to-moderate separations. Both patterns are consistent across the evaluated methods, indicating that they originate from the acquisition geometry rather than from a particular~matcher.

Fine-tuning on MOVR-SAR leads to three clear improvements. First, both RoMa v2 and FlexGlue achieve substantially higher accuracy across all angle bins and both platforms. Second, their accuracy varies only slightly with $\Delta\theta$, and performance in the opposite-direction regime becomes comparable to that in the same-direction regime, indicating a marked reduction in angle- and direction-dependent degradation. Third, FlexGlue outperforms RoMa v2 at essentially every separation, demonstrating greater robustness under the evaluated conditions. The degradation observed with official weights is therefore largely mitigated by target-domain training, although small residual variations remain.

\subsection{Limitations and Future Work}\label{sec:limitations}

This study has three main limitations, each of which also points to future work. First, the evaluation covers only Capella, Umbra, Sentinel-1, and NISAR; extending the benchmark to additional SAR platforms is important for assessing cross-platform generalization, although suitable multi-acquisition imagery is difficult to obtain from public sources. Second, registration between single-look complex (SLC) images remains to be validated, as the present results are not guaranteed to transfer to SLC data that preserve slant-range geometry and complex-valued phase; future work will extend both the benchmark and its evaluation to SLC imagery. Third, downstream applications remain underexplored; for example, a rigorous evaluation of SAR--SAR radiometric harmonization requires 16-bit imagery that is difficult to obtain consistently across platforms. In addition, the current subsets consist of intra-platform pairs; extending the benchmark to cross-band and cross-platform matching, for example between X-, C-, and L-band acquisitions, is a natural and important extension. More broadly, moving from registration toward downstream tasks such as change detection and three-dimensional reconstruction under cross-geometry acquisitions is a natural next step.

% =====================================================================
% ch6/ch6.tex
% Section 6: Conclusion
% =====================================================================

\section{Conclusion}\label{sec:conclusion}

This paper presented MOVR-SAR, a benchmark for SAR--SAR registration under varying acquisition geometries. Its unified framework combines four imaging-geometry pair categories with viewing-angle-separation labels, while a common coarse-to-fine, quality-controlled pipeline produces approximately 450\,000 registered and quality-screened tile pairs from Capella, Umbra, Sentinel-1, and NISAR. Spanning the X, C, and L bands, ground sampling distances of 0.25--20 m, and diverse land-cover types, MOVR-SAR supports geometry-aware training and stratified evaluation.

Experiments show poor generalization of official and cross-modal weights, especially on high-resolution X-band imagery. Fine-tuning on MOVR-SAR improves accuracy and reduces disparities across platforms, geometries, and viewing angles, demonstrating the importance of target-domain coverage. The same-orbit, same-look-side advantage is therefore not invariant, while land-cover effects remain platform-dependent and require stratified evaluation. MOVR-SAR provides a foundation for robust SAR--SAR registration and future extensions to additional platforms, SLC and cross-platform pairs, and downstream applications.\looseness=-1

\printcredits

\section*{Declaration of competing interest}

The authors declare that they have no known competing financial interests or personal relationships that could have appeared to influence the work reported in this paper.

\section*{Acknowledgments}

\begin{sloppypar}
This work was supported in part by the Xiong’an New Area Science and Technology Innovation Special Program under Grant 2025XAGG0042, in part by the National Natural Science Foundation of China (NSFC) under Grant 42671575 and Grant 42271444, and in part by the Postdoctoral Fellowship Program of CPSF under Grant Number GZB20250065, and in part by the China Postdoctoral Science Foundation under Grant Number 2026M790232.
\end{sloppypar}

%% Load the bibliography style file.
\bibliographystyle{elsarticle-num}
% Load the bibliography database.
\bibliography{cas-refs.bib}

\end{document}